\documentclass[lettersize,journal]{IEEEtran}
\usepackage{amsmath,amsfonts}
\usepackage{algorithmic}
\usepackage{algorithm}
\usepackage{array}
\usepackage{booktabs}
\usepackage[caption=false,font=normalsize,labelfont=sf,textfont=sf]{subfig}
\usepackage{textcomp}
\usepackage{stfloats}
\usepackage{float}
\usepackage{url}
\usepackage{verbatim}
\usepackage{graphicx}
\usepackage{cite}
\usepackage{colortbl}
\usepackage[colorlinks=true, linkcolor=black, urlcolor=black, citecolor=black]{hyperref}
\usepackage[dvipsnames]{xcolor}
\definecolor{revise}{HTML}{2975A7}
\definecolor{mygray}{gray}{.9}

\begin{document}

\title{Learning Pyramid-structured Long-range Dependencies for 3D Human Pose Estimation}

\author{Mingjie Wei, Xuemei Xie,~\IEEEmembership{Senior Member,~IEEE}, Yutong Zhong, Guangming Shi,~\IEEEmembership{Fellow,~IEEE}
\thanks{Manuscript received 05 May 2024; revised 09 Jul 2024 and 26 Oct 2024; accepted 25 Nov 2024. Date of publication X; date of current version X. This work was supported in part by the Natural Science Foundation (NSF) of China under Grant 62293483 and in part by Pazhou LAB (Huangpu) under Grant 2022K0903. The Associate Editor coordinating the review of this manuscript and approving it for publication was Prof. Roger Zimmermann. (Corresponding author: Xuemei Xie)}
\thanks{Mingjie Wei and Yutong Zhong are with the School of Artificial Intelligence, Xidian University, Xi’an 710071, China (email: mjwei@stu.xidian.edu.cn; ytzhong@stu.xidian.edu.cn)}
\thanks{Xuemei Xie is with the School of Artificial Intelligence, Xidian University, and also with Pazhou LAB (Huangpu), Guangzhou 510555, China (email: xmxie@mail.xidian.edu.cn)}
\thanks{Guangming Shi is with the School of Artificial Intelligence, Xidian University, and also with Peng Cheng Laboratory, Shenzhen 518066, China. (email: gmshi@xidian.edu.cn)}
\thanks{This paper has supplementary downloadable material available at http://ieeexplore.ieee.org., provided by the author.}
\thanks{Digital Object Identifier 10.1109/TMM.2025.3535349}}

\markboth{IEEE TRANSACTIONS ON MULTIMEDIA,~Vol.~25, 2024}%
{Shell \MakeLowercase{\textit{et al.}}: A Sample Article Using IEEEtran.cls for IEEE Journals}



\maketitle

\begin{abstract}
Action coordination in human structure is indispensable for the spatial constraints of 2D joints to recover 3D pose. Usually, action coordination is represented as a long-range dependence among body parts. However, there are two main challenges in modeling long-range dependencies.
First, joints should not only be constrained by other individual joints but also be modulated by the body parts. Second, existing methods make networks deeper to learn dependencies between non-linked parts. They introduce uncorrelated noise and increase the model size. 
In this paper, we utilize a pyramid structure to better learn potential long-range dependencies. It can capture the correlation across joints and groups, which complements the context of the human sub-structure. In an effective cross-scale way, it captures the pyramid-structured long-range dependence. Specifically, we propose a novel Pyramid Graph Attention (PGA) module to capture long-range cross-scale dependencies. It concatenates information from various scales into a compact sequence, and then computes the correlation between scales in parallel. Combining PGA with graph convolution modules, we develop a Pyramid Graph Transformer (PGFormer) for 3D human pose estimation, which is a lightweight multi-scale transformer architecture. It encapsulates human sub-structures into self-attention by pooling. Extensive experiments show that our approach achieves lower error and smaller model size than state-of-the-art methods on Human3.6M and MPI-INF-3DHP datasets. The code is available at \url{https://github.com/MingjieWe/PGFormer}.
\end{abstract}


\section{Introduction}
\label{sec:intro}
\IEEEPARstart{3}{D} human pose estimation (HPE) predicts human joint locations in the camera coordinate system from monocular images. It is crucial for applications such as action recognition \cite{10025821,8853267,tu2023implicit}, pose tracking \cite{9034193}, virtual reality \cite{mehta2017vnect}, etc. By accurately and efficiently processing human pose from images, downstream tasks are facilitated by reducing costs and improving performance. To promote the efficiency of the entire real-time multimedia system, a lightweight approach \cite{zhou2024blockgcn, zou2021modulated, zhao2022graformer} should be considered. It can be easier to deploy, more flexible, and less complex. Additionally, some complex actions lack robustness, which limits the application of 3D pose estimation, as certain motion patterns are difficult to model. Mainstream approaches \cite{cai2019exploiting,martinez2017simple,zhao2019semantic,zou2021modulated,zhao2022graformer, gong2023diffpose,chen2023hdformer} now conduct 3D HPE via 2D-to-3D lifting, which detects 2D keypoints and then lifts them to 3D. However, this lifting remains challenging due to the monocular ill-posed nature. Particularly, a single 2D pose can correspond to multiple valid 3D body configurations. Researchers have partially addressed it by exploiting spatial constraints among body joints, yet the connection structure of the joints relationship is not well modeled.\par
\begin{figure}[t]
\begin{minipage}{1.0\linewidth}
	\centering
	\centerline{\includegraphics[width=\textwidth]{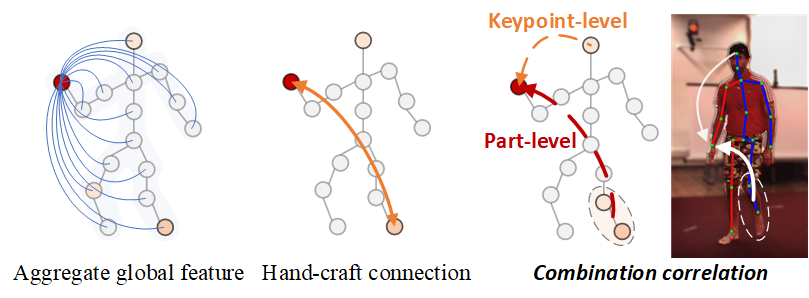}}
\end{minipage}
\begin{minipage}[b]{1.0\linewidth}
  \centering
  \centerline{\includegraphics[width=\textwidth]{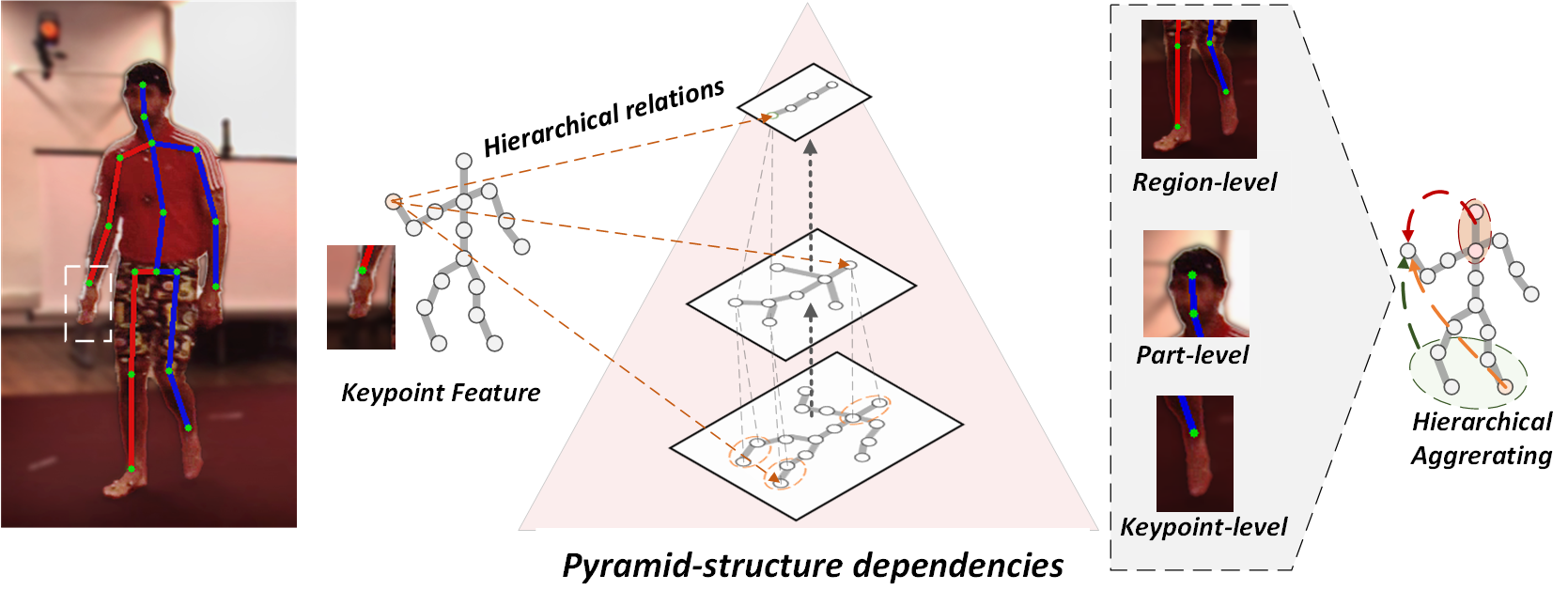}}
\end{minipage}
\caption{{\bf Top:} {Illustration of long-range dependence differences among global aggregation\cite{zou2021modulated, zheng20213d}, hand-crafted connections\cite{fang2018learning} and ours.} The proposed method learns potential correlation with fewer unrelated noise. {Joints should not only be constrained by other individual joints but also be modulated by the body parts.} {\bf Bottom:} We propose to capture long-range dependencies via pyramid structure. The relationship can be learned through cross-scale human sub-structure. {Hierarchical features facilitate the representation of complex motion dynamics.}}
\label{fig:illustrated}
\end{figure}
Most approaches \cite{cai2019exploiting,zhao2019semantic,xu2021graph,hu2021conditional,ci2019optimizing,liu2020comprehensive} focus on establishing short-range connections, by utilizing graph convolution networks (GCNs), to learn the reliable structure from graph data.
However, the relationships among body parts should not be confined to short-range interactions between neighboring joints. Some studies \cite{fang2018learning,zou2021modulated,zeng2021learning,zhao2022graformer} have revealed that certain complex actions exhibit extensive coordination beyond natural connections, e.g., the strong correlation between head and hand during eating, the coordination between hands and legs while walking, so are the two symmetrical legs. 
Thus, current methods consider that action coordination is expressed as a long-range dependence which is not restricted to first-order neighbors and natural skeletal connections. However, the long-range dependence of connection structure is not adequately modeled.
\par
{In capturing long-range dependencies, current work suffers from two limitations: most methods overlook part-level relationships, while others model human sub-structure by increasing the model size.}\par 
{ {\it First}, Fang et al. \cite{fang2018learning} hand-craft the connections among long-range joints, overlooking {\it potential} long-range dependencies by focusing only on specific node pairs. Zou et al. \cite{zou2021modulated} appended a learnable long-range matrix to the skeletal graph convolution. This is constrained by primitive skeletal limitations and lacks extensive long-range dependency modeling. In contrast, spatial self-attention effectively assigns adaptive weights to long-range keypoints \cite{zheng20213d, zhao2022graformer}. But it introduces unrelated noise. Thus, it necessitates an enhanced connectivity structure to  effectively modeling distant relations. This means modeling human structures prior in self-attention rather than learning purely from pixel positions.}\par
{So, how can attention mechanisms be efficiently utilized to model long-range dependencies? Notably, existing approaches ignore {\it potential part-level} relations. As shown in Fig. \ref{fig:illustrated}, each body joint coordinates with different hierarchical connections rather than with individual pairwise joints, as modeled in current methods  \cite{zou2021modulated, fang2018learning,zhao2022graformer,zheng20213d}. Thus, we propose combining different hierarchical correlation to accurately predict 3D pose keypoints. Because such dependencies naturally exist, as seen in the coordination between hands and feet and the synergy of muscle groups.}\par
{\it Second}, some methods expand the receptive field by increasing the depth \cite{xu2021graph} or branches \cite{zhang2023learning}, aiming to learn relationships among parts. {We discover that multi-scale information can model the human body sub-structure. Hierarchical  long-range constraints help model complex human actions, such as sitting, which tend to have higher error rates.} However, aggregating information from all scales introduces noise and increases the model size.\par
To solve the problems mentioned above, we propose to encapsulate human sub-structures into self-attention by pooling. We design a pyramid structure for feature fusion and use self-attention to compute cross-scale correlations. This not only effciently models the relationship between different scales, but also achieves the modeling of long-range dependencies.\par
{In conclusion, in this paper, we propose to capture long-range dependencies through a simple and efficient approach from a multi-scale perspective. Specifically, we propose to encapsulate human sub-structures into self-attention via a pyramid structure, termed Pyramid Graph Attention (PGA) module. The proposed pyramid structure provides these different hierarchical connections, encompassing keypoints, limb parts, and body regions, enabling a representation of complex motion dynamics. It extracts long-range semantic features from cross-scale information and computes the correlation between the original and pooled scales in parallel. Furthermore, we introduce a novel multi-scale architecture for 3D HPE, named Pyramid Graph Transformer (PGFormer). We introduce a pyramid structure in the attention mechanism, rather than learning each scale independently \cite{xu2021graph,zhang2023learning}, making the architecture simple and lightweight. We prove the universality of our PGA module through plug-and-play experiments, and design a diffusion based model called DiffPyramid. {This hierarchical spatial constraint relationship is necessary not only for direct modeling methods but also for distribution-based methods \cite{gong2023diffpose} to effectively model joint-wise relations. It uses PGFormer to initialize 3D Pose and achieves better results. It is further demonstrated that hierarchical spatial information of long-distance constraints can effectively improve the diffusion model performance.}} \par
Extensive experiments show that our approach achieves smaller error and smaller model size than state-of-the-art method on Human3.6M and MPI-INF-3DHP datasets. {This means that PGFormer can learn efficient feature representations with a small number of parameters and fast inference speed. Additionally, we carry out visualizations to prove the robustness of PGFormer. Through extensive experiments, we demonstrate that PGFormer has learned long distance dependencies and achieved significant performance improvements.} \par
\begin{figure}[t]
\begin{minipage}[b]{1.0\linewidth}
  \centering
  \centerline{\includegraphics[width=9cm]{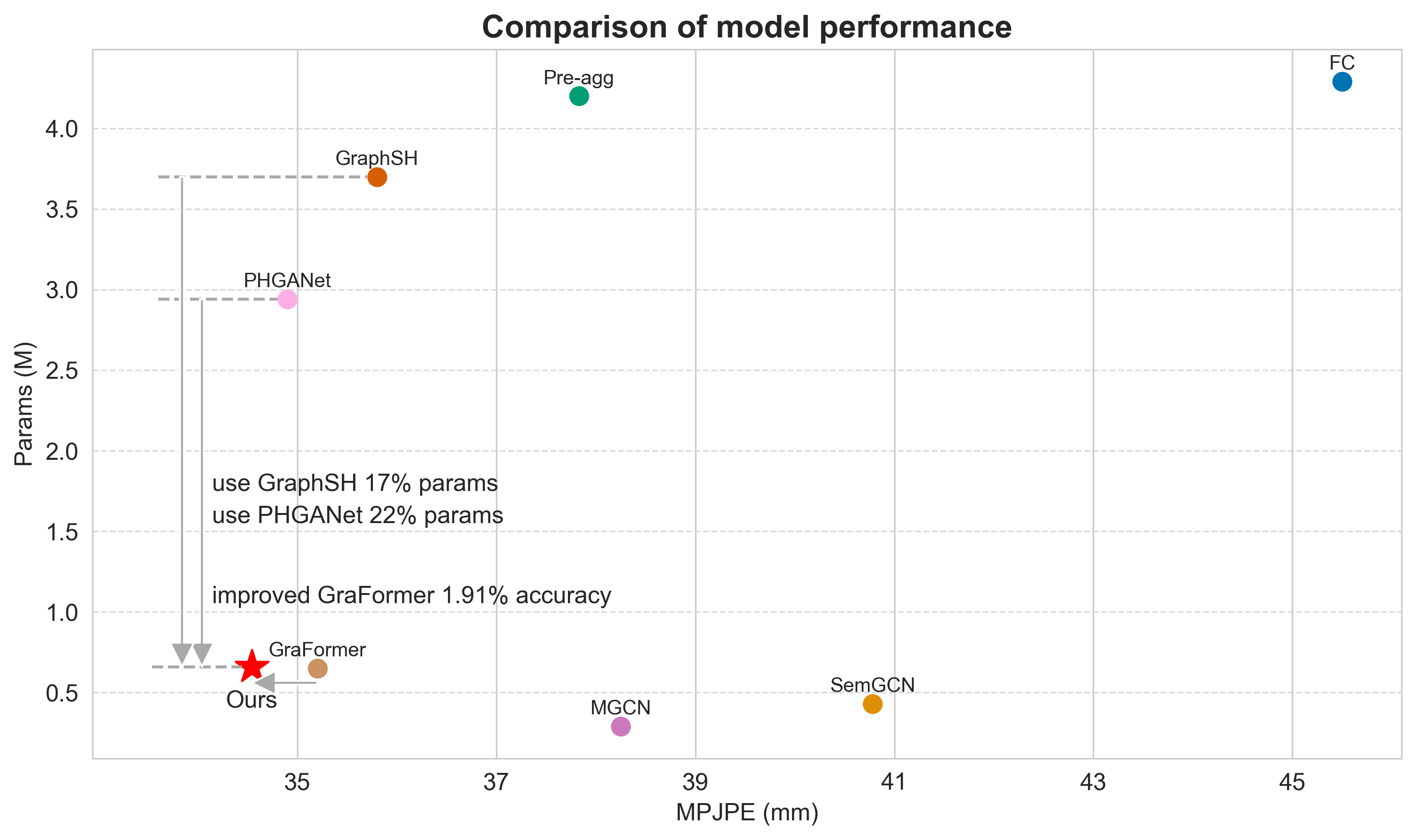}}
  \medskip
\end{minipage}
\caption {Comparison of the performance and model size between the proposed PGFormer and state-of-the-art  on Human3.6M \cite{ionescu2013human3} with ground truth of 2D joints as input. Compared with baseline GraFormer\cite{zhao2022graformer}, PGFormer reduces 1.9\% MPJPE error. Compared with the multi-scale method, e.g. GraphSH \cite{xu2021graph} and PHGANet \cite{zhang2023learning}, our parameters have been reduced by 83\% and 78\%, respectively. }
\label{fig:ComprehensivePerformanceShow}
\end{figure}

In sum, our work makes the following contributions:
\begin{itemize}
\setlength{\itemsep}{0pt}
\setlength{\parsep}{0pt}
\setlength{\parskip}{0pt}
\item[$\bullet$] The proposed Pyramid Graph Attention (PGA) module learns long-range dependencies by capturing correlations across joints and body parts, yielding reliable results.
\item[$\bullet$] The proposed Pyramid Graph Transformer (PGFormer) is a {\it simple} and {\it lightweight} architecture for extracting hierarchical features to perform 3D HPE.
\item[$\bullet$] {The designed PGA module and PGFormer exhibit strong universality and scalability. Plug-and-play experiments and DiffPyramid demonstrate that they can be applied to both GCN-based and diffusion-based architectures.}
\item[$\bullet$] Compared with state-of-the-art methods, our architecture improves accuracy and reduces model size on the Human3.6M dataset, as shown in Fig. \ref{fig:ComprehensivePerformanceShow}. It also demonstrates better generalization on the MPI-INF-3DHP dataset.
\end{itemize}

\section{Related Work}
\label{sec:relatedwork}
\subsection{3D Human pose estimation}
{The inference of human body coordinates in 3D space from a single image was first proposed by Lee and Chen \cite{lee1985determination}. Recently, state-of-the-art approaches have employed deep neural networks. Some methods use end-to-end regression \cite{mehta2017monocular, pavlakos2017coarse} to predict 3D coordinates or heatmaps directly from a single image. For instance, Pavlakos et al. \cite{pavlakos2017coarse} proposed a coarse-to-fine network that predicts depth heatmaps using a convolutional neural network. However, these methods struggle with the mapping from a 2D image to a 3D human body.}\par
{With advancements in 2D HPE, researchers decoupled the problem of 3D HPE and addressed it through 2D-to-3D lifting. \cite{cai2019exploiting,martinez2017simple,zhao2019semantic,zou2021modulated,zhao2022graformer,ZhongTMM2024,WangTMM2024,tang20233d,li2022mhformer, liu2023posynda}. This approach is capable of exploring spatial \cite{cai2019exploiting,martinez2017simple,zhao2019semantic,zou2021modulated,zhao2022graformer} and temporal \cite{ZhongTMM2024,WangTMM2024,tang20233d,li2022mhformer, chen2023hdformer} information to achieve excellent performance,  Consequently, we adopt this two-stage approach as well.
To model human structure and exploit the relations, some works \cite{cai2019exploiting,zhao2019semantic,hu2021conditional,ci2019optimizing,liu2020comprehensive} are based on GCN network architecture. For example, Zhao et al. \cite{zhao2019semantic} use semmantic graph convolution and non-local modules \cite{wang2018non} to learn spatial constrains. However, local graph convolution aggregates information from adjacent keypoints but lacks the establishment of long-range coordination. The method we proposed builds upon local information and delves into the modeling of long-range dependencies.}\par 
In addition, some distribution-based methods \cite{gong2023diffpose, holmquist2023diffpose, liu2023posynda} propose to learn pose distribution. {For instance, a reliable 3D pose is estimated by using diffusion model \cite{gong2023diffpose}. But they also require a conditional reverse diffusion step by modeling the spatial context. The proposed hierarchical long-range dependencies are also helpful to capture spatial priors and model sub-structures at different levels in the diffusion process.} \par
\subsection{Long-range dependence}
Long-range dependence refers to the dependencies among non-adjacent nodes in 3D HPE \cite{zou2021modulated}. In recent years, some methods \cite{fang2018learning, he2021db,zeng2021learning,zhao2022graformer} have recognized this widespread dependence in the structure of human body. It is often used to model human coordination in different actions. Therefore, taking this into consideration yields better results in some complex actions.\par
{Fang et al. \cite{fang2018learning} propose to learn the symmetry and coordination of specific joint pairs via hand-craft connection. Zou and Tang \cite{zou2021modulated} propose to learn the motion patterns beyond natural connection via aggregating all high-order nodes. However, this manual design and reliance on graph convolution limits the efficiency of learning long-range constraints. }
Nowadays, some of the latest methods \cite{zhao2022graformer,zhang2023learning,li2022exploiting,li2022mhformer,zhang2022mixste} design the network architecture based on attention  \cite{vaswani2017attention}. For example, Zhao et al. \cite{zhao2022graformer} design the transformer architecture that combines graph convolution and attention. Li et al. \cite{li2022exploiting} use transformer to exploit long-range dependence. {However, calculation process of self-attention ignores the rich structural information of human body. Previous work \cite{zhao2022graformer} improves long-range dependencies learning by replacing MLP with convolution, and does not improve the fundamental problem of the lack of structural information in attention. So, our approach further considers improving the self-attention mechanism to learn efficient feature representations of long distance dependencies, rather than noise data due to pure coordinate values.}
\subsection{Hierarchical human structure}
{The hierarchical human structure is a critical concept in the field of human pose estimation. It involves modeling the human body as a series of interconnected parts with different levels of hierarchy. This representation reflects the natural organization of the human body and its joints, allowing for more accurate and realistic predictions of poses. Grouping \cite{zhou2019hemlets,zeng2020srnet,xue2022boosting,wu2022hpgcn} or pooling\cite{xu2021graph, hua2022unet, zhang2023learning} are used to achieve hierarchical representation, such as the proposed pyramid structure. Different from image tasks \cite{wu2022p2t,PVT}, in 3D HPE, the pyramid structure provides graph-like hierarchical information about the substructure of the human body.}\par
{These methods\cite{zhou2019hemlets,zeng2020srnet,xue2022boosting,wu2022hpgcn} demonstrate the importance of part-level analysis in 3D HPE. For example, Zhou et al. \cite{zhou2019hemlets} propose to learn part-centric heatmaps and Wu et al.\cite{wu2022hpgcn} represented human structure using hierarchical poselets. However, they do not adopt a explicit pooling approach to directly acquire part-level information. The proposed pyramid structure facilitates adequate information exchange across multiple scales. And it is an intuitive and efficient method to extract human substructure feature by pooling, which is explainable.
Similarly, Xu et al. \cite{xu2021graph} construct a Graph Hourglass network by pooling on the hierarchical representation of the human skeleton. The proposed pyramid retains the original scale information, and achieves cross-scale calculation, which is conducive to the utilization of multi-scale information. Additionally, Zhang et al. \cite{zhang2023learning} introduce a parallel framework to compute semantic relations between different scales for 3D human pose estimation, but the architecture is redundant. We propose to integrate multi-scale concepts into self-attention, which can learn human substructure features in parallel and calculate correlations using a small number of parameters. In sum, we implement pooling in the self-attention to efficiently learn hierarchical human structure information.} \par

\begin{figure*}[t]
\begin{minipage}[b]{1.0\linewidth}
  \centering
  \centerline{\includegraphics[width=18cm]{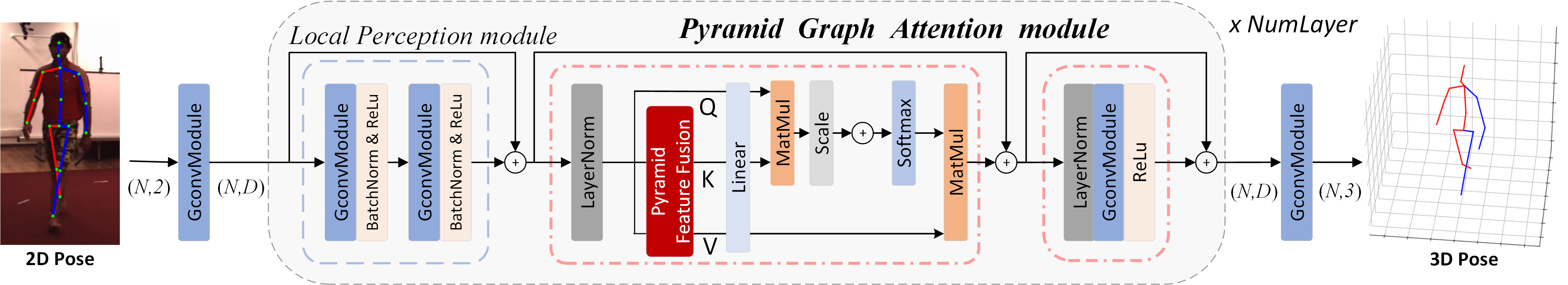}}
  \medskip
\end{minipage}
\caption {The network architecture of the proposed {\bf PGFormer} for 3D human pose estimation. The PGFormer consists of a Local Perception module and a Pyramid Graph Attention module. {\bf Local Perception modules} are stacked by graph convolution layers.  {\bf Pyramid Graph Attention module} consists of a multi-head attention with {\it pyramid fusion} (red) and a graph convolution layer. ($N,D$) donates  the number of 2D joints and feature channels, respectively.}
\label{fig:network architecture}
\end{figure*}

\section{Our Approach}
\subsection{Overview}
As illustrated in Fig. \ref{fig:network architecture}, the proposed PGFormer can be input by 2D pose keypoints obtained from a 2D detector and predict 3D pose in a single frame. It is built upon stacking two main modules: Local Perception module and Pyramid Graph Attention module. The Local Perception module is formed by interleaving graph convolution layers, batch norm and ReLU. It is presented alongside vanilla GCN in Sec. \ref{subsection:GCN}. Introduced in Sec. \ref{subsection:PGAM}, the proposed Pyramid Graph Attention module contains a special pyramid multi-head attention and a graph convolution layer. Each module of our architecture we have mentioned is connected by residuals.
\subsection{GCN Layers}
\label{subsection:GCN}
Graph convolution in GCN blocks is used to model physical graph-like features of the human skeleton. A graph can be defined as $G=\{V,E\}$, where $V$ is the set of nodes $N$ and $E$ is the set of edges. The edges can be represented by an adjacency matrix $A\in[0,1]^{N \times N}$. Node representations in $l$-th layer of a GCN are collected into a matrix $H^l \in \mathbb{R}^{d_1 \times N}$ and transformed
by a learnable parameter matrix $W \in \mathbb{R}^{d_2 \times d_1}$, where $d_1$ and $d_2$ are the dimensions before and after the transformation, respectively. The formula of the vanilla GCNs is as follows:
\begin{equation}
H^{(l+1)}=Gconv(H)=\sigma(WH^l\tilde{A}),
\end{equation}
where $H^{(l+1)} \in \mathbb{R}^{d_2 \times N}$ is the updated feature matrix, $\sigma$ is
the activation function like ReLU \cite{nair2010rectified}, and $\tilde{A} = I - D^{-1/2}AD^{-1/2}$ is the normalized adjacency matrix.\\
{\it Local Perception module} aggregates features of the human skeleton. In our design, Local Perception module is a replaceable module. So different graph convolution can be used, e.g SemGCN \cite{zhao2019semantic} and Modulated GCN \cite{zou2021modulated}. In this module, we adopt Chebyshev graph convolution \cite{defferrard2016convolutional} due to its more powerful handling of graph data. The formula of Chebyshev graph convolution is:
\begin{equation}
H^{(l+1)}=\sum_{k=0}^{K-1} T_k \tilde{L} H^l W_k,
\end{equation}
where $T_k(x)=2xT_{k-1}(x)-T_{k-2}(x)$ represents the Chebyshev polynomial of degree $k$,
and $T_0=0$, $T_1=x$, $\tilde{L}=2\tilde{A}\/\lambda-I$, $\lambda$ is the maximum eigenvalue of $\tilde{A}$. The graph convolution is able to aggregate $K$-top neighbors information of a joint.
\begin{figure}[t]
\begin{minipage}[b]{1.0\linewidth}
  \centering
  \centerline{\includegraphics[width=9cm]{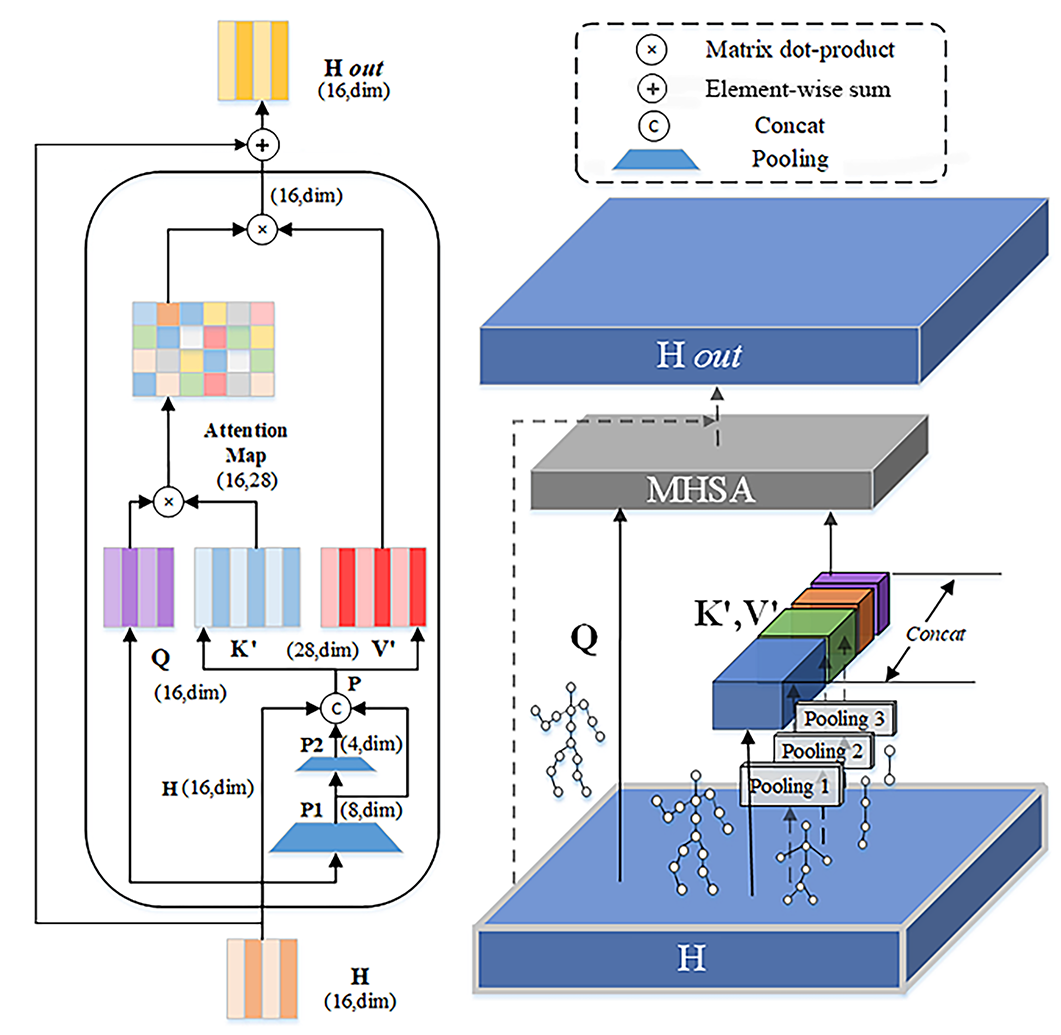}}
  \medskip
\end{minipage}
\caption {Architecture and illustration of Pyramid Graph Attention. {We construct the pyramid body structure using the input $H$, and attention captures cross-scale long-range dependencies.} {\bf Left:} The figure introduces the specific calculation and dimensional change in MHSA. {\bf Right:} The figure is a schematic of the pyramid concatenation in the proposed method. }
\label{fig:SelfAttentionShow}
\end{figure}
\subsection{Pyramid Graph Attention module}
\label{subsection:PGAM}
{Pyramid Graph Attention module consists of a multi-head attention with pyramid fusion and a graph convolution layer, as shown in Fig. \ref{fig:network architecture}. Both the multi-head attention and the graph convolution layer are equipped with residual connections.}\par
{\it Pyramid Feature Fusion} is a unique part of the proposed Pyramid Graph Attention module as shown in Fig. \ref{fig:SelfAttentionShow}. Pyramid information of different resolutions gathers via fusion operation which is beneficial for subsequent parallel computation. We adopt the graph pooling strategy proposed in GraphSH \cite{xu2021graph} to achieve transformation across different scales. Extensively, we can use different group pooling methods to generate pyramid feature maps, like:
\begin{equation}
\begin{aligned}
P_1&=AvgPooling_1 (H),\\
P_2&=AvgPooling_2 (H),\\
&\cdots\\
P_n&=AvgPooling_n (H),\\
\end{aligned}
\label{equation:Pooling}
\end{equation}
where $\{P_1, P_2, ..., P_n\}$ constitute the pyramid feature map and $n$ is the number of different pooling layers. {Specifically, we use average pooling via manual groups. We employ two pooling operations to construct a pyramid structure for achieving optimal results. $P_1$  represents the part-level features via $AvgPooling_1(H)$ from keypoint-level features. The pooling layer aggregates the features from two nodes. $P_2$  indicates region-level features from $AvgPooling_2(H)$. It aggregates the features from four nodes. This $P_2$ feature contains information about the hands, legs, torso, and head. Then, the pyramid feature $P$ can be derived through layer normalization and features concatenation, which can be formulated as:}
\begin{equation}
P=LayerNorm(Concat(H,P_1,P_2)),
\label{equation:PoolingConcat}
\end{equation}
where the matrix $H$ in original scale is retained and appends information on multiple scales. {Exactly, the fusion process includes original scale features $H$, body parts features $P_1$and body regions features $P_2$.}\par
{\it Multi-head Self Attention (MHSA)} layer learns long-range dependencies beyond the human skeleton. The correlation between original and other scales is calculated in a parallel manner through MHSA. Feature representation is used as input, then projected to
query $Q\in \mathbb{R}^{N \times D}$, key $K\in \mathbb{R}^{N \times D}$, and value $V \in \mathbb{R}^{N \times D}$:\\
\begin{equation}
Q=HW_Q,K=HW_K,V=HW_V.
\end{equation}
We use pyramid features $P$ instead of $H$ to get the converted key $K'$ and value $V'$ after pyramid feature fusion, which can be formulated as follows:
\begin{equation}
Q=HW_Q,K'=PW_K,V'=PW_V.
\end{equation}
Multiple headers are used for joint modeling of information on different representations of feature sub-spaces. The output of MHSA
is a concatenation of $h$ attention head outputs as formulated:
\begin{equation}
X_i = \sigma(\frac{{Q}{K'}^T}{{\sqrt{D}}})V',i\in(1,...,h),
\end{equation}
\begin{equation}
MHSA(Q,K,V) = Concat(X_1, X_2, ..., X_h)W_{out},
\end{equation}
where $X_i$ is a scaled dot-product attention receives $Q$, $K'$, and $V'$ , {$\sigma(\cdot)$ is a softmax function. Subsequently, a dropout layer is employed following the softmax operation to augment regularization effectiveness.} Then, scaled dot-product attention is applied by each head $i$ in parallel. {As shown in Fig. \ref{fig:SelfAttentionShow}, a shortcut connection makes the model easier to optimize. Therefore, the result $H_{out}$ can be formulated as:}
\begin{equation}
	H_{out} = H + MHSA(Q,K,V).
\end{equation}
{As shown in Fig. \ref{fig:network architecture}, a graph convolution is used to process graph data after self-attention. Clearly, we use a vanilla graph convolution and ReLU to replace the MLP layers from the standard transformer. A complete layer result $F$ can be formulated as:}
\begin{equation}
F = H_{out} + ReLu(Gconv(H)),
\end{equation}
{where $H_{out}$ refers to the original output after self-attention, $H$ represents the features after layer normalization, and $Gconv(\cdot)$ denotes the vanilla graph convolution operation. In order to make the whole model more robust, a dropout is set after self-attention and graph convolution. Finally, the pyramid-structured multi-head attention models long-range dependencies, correlating hierarchical features with keypoint features for a comprehensive representation. The PGA module thoroughly learns the prior knowledge of human body structure.}\par
\subsection{Pyramid Diffusion Model}
\begin{figure}[htb]
	\begin{minipage}[b]{1.0\linewidth}
		\centering
		\centerline{\includegraphics[width=8cm]{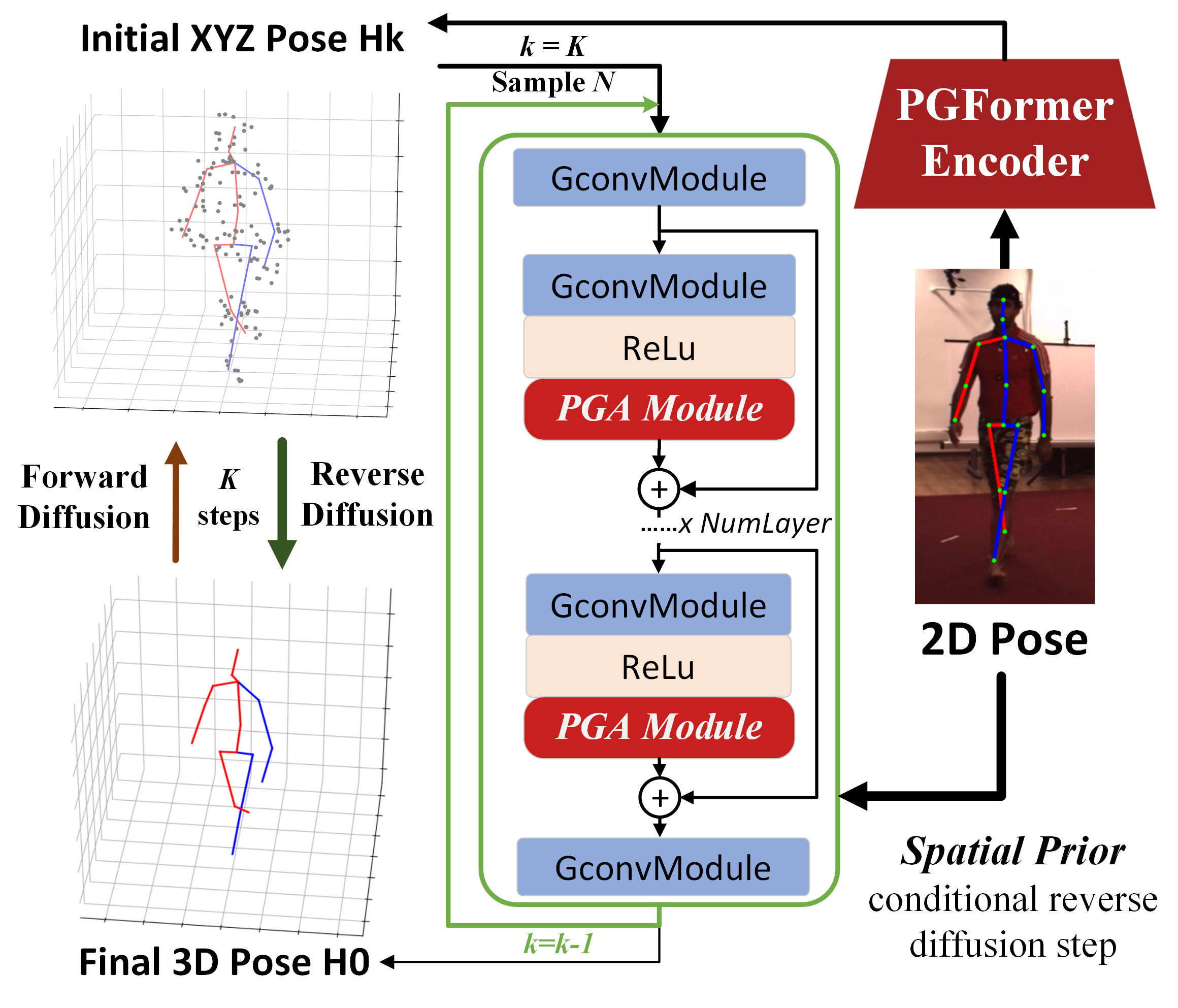}}
		\medskip
	\end{minipage}
	\caption {Architecture of the proposed DiffPyramid for 3D HPE. {On the left, it features two processes: the forward noise addition \(\epsilon\) during training and the reverse denoising process during inference. In the middle, the PGFormer network serves as the diffusion model \( g \), capturing joint-wise relationships. Its PGA module incorporates prior knowledge of the human body sub-structure. On the right, 2D pose as a spatial prior to guide the reverse diffusion steps and a pre-trained PGFormer provides an initial \( z \) value to initialize \( H_k \).}}
	\label{fig:DiffPyramid}
\end{figure}

{We further designed a two-step diffusion model, refer to \cite{gong2023diffpose, holmquist2023diffpose}. As illustrated in Fig. \ref{fig:DiffPyramid}, it consists of two processes: the forward process and the reverse process. In the forward process, the model is supervised during training by adding $K$ times noise $\epsilon$ to the deterministic 3D ground truth $H_0$. In the reverse process, the input is the initialized indeterminate 3D pose distribution $H_K$, which is denoised using the diffusion model $g$ (parameterized by $\theta$). We use the contextual features of the 2D pose, obtained through PGFormer, as conditions to predict the 3D pose. In other words, it achieves 3D human pose estimation through a reverse diffusion process, converting a noisy and uncertain pose into a definite pose. }\par
{Specifically, during inference, DiffPyramid is constructed by using the network architecture as Fig. \ref{fig:DiffPyramid} which is stacked by graph convolution and PGA modules. The inference consists of two stages: initializing the indeterminate 3D pose $H_k$ and the reverse diffusion process via diffusion model $g$.}\par
{{\it Initializing Condition 3D Pose Distribution.} To construct the 3D pose distribution, the process involves constructing the uncertainty distribution of the x, y, and z coordinates for each 2D pose joint. According to \cite{gong2023diffpose, holmquist2023diffpose}, an initial distribution with sample-specific knowledge is more suitable for solving 3D HPE, rather than Gaussian noise without any prior in vanilla diffusion model \cite{ho2020denoising,ddim}. Thus, the $x$ and $y$ information of indeterminate pose distribution $H_k$ is initialized by given 2D pose, and then the {\it z} information is obtained by PGFormer as an encoder to obtain the initial pose distribution. We use a Gaussian Mixture Model (GMM) in \cite{gong2023diffpose} to model the uncertainty distribution $H_k$ for 3D pose estimation.}\par
{The next step is to gradually reduce the uncertainty from $H_k$ to obtain $H_0$ using the diffusion model $g$, i.e, do {\it reverse diffusion}. $N$ noise poses $\{{h_k^i}\}_{i=1}^N$ are sampled from $H_k$. The process in vanilla diffusion model \cite{ho2020denoising,ddim} of transforming an inaccurate sample $h_k$ from $N(0,I)$ into an accurate sample $h_0$ through repeated denoising is called {\it reverse diffusion} ($h_k->h_{k-1}->...->h_0$). Conversely, the process from $h_0$ to $h_k$ is referred to as {\it forward diffusion}. To enable the model to have denoising capabilities, we first need to perform forward diffusion. It utilizes ground truth for supervised training by adding noise $\epsilon$.}\par
{{\it Forward Diffusion Progress} is a Markov chain that gradually adds noise parameterized by variance $\beta_k$  to \( h_0 \) at each time step \( k \), ultimately transforming \( h_0 \) into a noisy distribution $h_k\sim \mathcal{N}(0, I)$ that approximates the posterior \( q(h_{1:k}|h_0) \). It can be formoulated as:}
\begin{equation}
	q(h_k|h_{k-1}) := \mathcal{N}(h_k; \sqrt{1-\beta_k}h_{k-1},\beta_kI),
\end{equation}
\begin{equation}
	q(h_{1:k}|h_{0}) := \prod_{k=1}^{K}q(h_k|h_{k-1}).
\end{equation}\par
{The forward process allows for sampling $h_k$ at any time step in closed form as follows:}
\begin{equation}
 	q(h_{k}|h_{0}) := \mathcal{N}(h_k;\sqrt{\bar{\alpha}_k} h_0,(1-\alpha_k)I),
\end{equation}
{where $\alpha_k := 1-\beta_k$ and $\bar{\alpha}_k := \prod_{s=1}^{k}\alpha_s $.}\par 
{{\it Reverse diffusion}  is defined as the joint distribution $p_\theta(h_{0:K})$, which involves learned Gaussian transitions starting from  $p(h_K)=\mathcal{N}(h_K;0, I)$, as follow:}
\begin{equation}
	p_\theta(h_{k-1}|h_k):=\mathcal{N}(h_{k-1};\mu_\theta(h_k,f_k,f_c), \sum\nolimits_{\theta}(h_k,k)),
\end{equation}
{where we follow DDIM \cite{ddim} to set $\sum\nolimits_{\theta}(h_k,k)=\sigma^2I$ and }
\begin{equation}
	\mu_\theta(h_k,f_k,f_c) = \frac{1}{\sqrt{\alpha_k}}(h_k-\frac{\beta_k}{\sqrt{1-\bar{\alpha}_k}} \epsilon_\theta(h_k,f_k,f_c)),
\end{equation}
{the $f_c$ is condition feature in each diffusion step, specifically, we use the x and y of 2D Pose as the condition; the $f_k$ represents the unique step embedding feature at the $k^{th}$ step, obtained via a sinusoidal function. Thus, the predicted noise $\epsilon$ is conditioned on $f_c$ from $h_k$.
The reverse diffusion step can simply be expressed with a trained diffusion model $g$ and its parameters $\theta$, as follow:}
\begin{equation}
	h_{k-1}=g_{\theta}(h_k, f_k, f_c), k \in \{1,...,K\}.
\end{equation}\par
{Finally, the each sample $\{{h_0^i}\}_{i=1}^N$ represents the determined pose distribution $H_0$ obtained through denoising. The estimated 3D pose is obtained by computing the mean $\{{h_0^i}\}_{i=1}^N$.}
\par
\vspace{0.5em}
{{\it Relationship modeling in diffusion model.} We adopted the architecture of PGFormer for $g$ to achieve 3D pose estimation. GCN Layers (Local Perception modules in PGFormer) model the basic topological relationships between keypoints. Then, the proposed PGA module further learns the relationship between distant nodes and captures cross-scale information at different levels.}\par
{In short, our methods (PGA module and PGFormer) have strong universality and extensibility, which is applicable not only to GCN-based methods, but also to diffusion models. The stronger spatial prior information provided by PGFormer and the hierarchical long-distance dependencies provided by PGA module improve the ability of modeling human body structure.}\par
\subsection{Training}
PGFormer aims to establish 2D to 3D pose mapping when training. To be specific, Mean Squared Error (MSE) is employed as a loss function during training, which minimizes the errors between predicted pose and ground truth. In addition, we employ $l_1$-norm regularization to enhance sparsity in the loss function when employing various graph convolutions in the Local Perception block. So, we design the loss function $L$ to be implemented by a weighted summation:
\begin{equation}
	L = \lambda\sum_{i=1}^{N}\left\|   Y_i-\tilde{Y_i}  \right\|_1+\
	(1-\lambda)\sum_{i=1}^{N}\left\|   Y_i-\tilde{Y_i}  \right\|_2,
\end{equation}
where $\lambda$ is a  parameter of weight. $\left\| * \right\|_1$ indicates the $l_1$-norm. $\left\| * \right\|_2$ indicates the MSE. $N$ is the number of joints.\par 
{PGFormer Encoder in DiffPyramid has same configuration, but the diffusion model $g$ is different. During training, $H_k$ is initialized using PGFormer. Supervision for each step is derived from the forward diffusion process based on $H_k$ and the ground truth $H_0$. Each sample $i$ optimizes the model parameters $\theta$ progressively from $h_k^i$ to $h_{k-1}^i$ in a step-wise manner, and all $h_0^i = h_0$. Following \cite{gong2023diffpose, ho2020denoising,ddim}, the loss $L$ can be formulaated as:}
\begin{equation}
	L = \sum_{i=1}^N\sum_{j=1}^K \left\|  g_\theta(h_k^i f_k^j, f_c) - h_{k-1}^i \right\|_2^2.
\end{equation}

\section{Experiments}
\begin{table*}[t]
	\caption{Results comparison with all state-of-the-art methods on Human3.6M under MPJPE (mm). CPN detections 2D keypoints are used as input. No pose refinement applied in post-processing. The best and the second-best scores of single-frame methods are shown in bold and underlined, respectively. (†) uses temporal information.}
	\resizebox{2\columnwidth}{!}{
		
		\begin{tabular}{*{17}{l}}
			\toprule
			Method
			&Dir.&Disc&Eat&Greet&Phone&Photo&Pose&Purch.&Sit&SitD.&Smoke&Wait&WalkD.&Walk&WalkT.
			&Avg ↓\\
			\midrule
			Lee et al.\cite{lee1985determination} ECCV’18 (†) & 40.2  & 49.2  & 47.8  & 52.6  & 50.1  & 75.0  & 50.2  & 43.0  & 55.8  & 73.9  & 54.1  & 55.6  & 58.2  & 43.3  & 43.3  & 52.8  \\ 
			Cai et al.\cite{cai2019exploiting} ICCV’19 (†)  & 44.6  & 47.4  & 45.6  & 48.8  & 50.8  & 59.0  & 47.2  & 43.9  & 57.9  & 61.9  & 49.7  & 46.6  & 51.3  & 37.1  & 39.4  & 48.8  \\ 
			Pavllo et al.\cite{pavllo:videopose3d:2019} CVPR’19 (†) & 45.2  & 46.7  & 43.3  & 45.6  & 48.1  & 55.1  & 44.6  & 44.3  & 57.3  & 65.8  & 47.1  & 44.0  & 49.0  & 32.8  & 33.9  & 46.8  \\ 
			Wang et al.\cite{wang2020motion} ECCV'20(†) & 40.2  & 42.5  & 42.6  & 41.1  & 46.7  & 56.7  & 41.4  & 42.3  & 56.2  & 60.4  & 46.3  & 42.2  & 46.2  & 31.7  & 31.0  & 44.5  \\ 
			Zheng et al.\cite{zheng20213d} ICCV'21(†) & 41.5  & 44.8  & 39.8  & 42.5  & 46.5  & 51.6  & 42.1  & 42.0  & 53.3  & 60.7  & 45.5  & 43.3  & 46.1  & 31.8  & 32.2  & 44.3  \\ 
			Li et al.\cite{li2022mhformer} CVPR'22(†) & 39.2 & 43.1 & 40.1 & 40.9 & 44.9 & 51.2 & 40.6 & 41.3 & 53.5 & 60.3 & 43.7 & 41.1 & 43.8 & 29.8 & 30.6 & 43.0 \\
			Li et al.\cite{li2022exploiting} TMM'22(†) & 40.3 & 43.3 & 40.2 & 42.3 & 45.6 & 52.3 & 41.8 & 40.5 & 55.9 & 60.6 & 44.2 & 43.0 & 44.2 & 30.0 & 30.2 & 43.7 \\
			Tang et al.\cite{tang20233d} CVPR'23(†) & 40.6 & 43.0 & 38.3 & 40.2 & 43.5 & 52.6 & 40.3 & 40.1 & 51.8 & 57.7 & 42.8 & 39.8 & 42.3 & 28.0 & 29.5 & 42.0\\
			Zhou et al.\cite{zhou2023diff3dhpe} ICCVW'23(†) & 40.2  & 42.7  & 38.6  & 40.8  & 42.6  & 50.0  & 40.3  & 40.2  & 52.5  & 55.1  & 43.6  & 41.3  & 42.9  & 29.5  & 29.5  & 42.0 \\
			\midrule
			Yang et al.\cite{yang20183d} CVPR'18 & 51.5  & 58.9  & 50.4  & 57.0  & 62.1  & 65.4  & 49.8  & 52.7  & 69.2  & 85.2  & 57.4  & 58.4  & {\bf 43.6}  & 60.1  & 47.7  & 58.6  \\ 
			Fang et al.\cite{fang2018learning} AAAI' 18 & 50.1  & 54.3  & 57.0  & 57.1  & 66.6  & 73.3  & 53.4  & 55.7  & 72.8  & 88.6  & 60.3  & 57.7  & 62.7  & 47.5  & 50.6  & 60.4  \\ 
			Zhao et al.\cite{zhao2019semantic} CVPR' 19 & 48.2  & 60.8  & 51.8  & 64.0  & 64.6  & {\bf 53.6}  & 51.1  & 67.4  & 88.7  & 57.7  & 73.2  & 65.6  & 48.9  & 64.8  & 51.9  & 60.8  \\ 
			Sharma et al.\cite{sharma2019monocular} ICCV’19 & 48.6  & 54.5  & 54.2  & 55.7  & 62.2  & 72.0  & 50.5  & 54.3  & 70.0  & 78.3  & 58.1  & 55.4  & 61.4  & 45.2  & 49.7  & 58.0  \\ 
			Liu et al.\cite{liu2020comprehensive} ECCV’20 & 46.3  & 52.2  & 47.3  & 50.7  & 55.5  & 67.1  & 49.2  & 46.0  & 60.4  & 71.1  & 51.5  & 50.1  & 54.5  & 40.3  & 43.7  & 52.4  \\ 
			Xu and Takano\cite{xu2021graph} CVPR' 21 & 45.2  & 49.9  & 47.5  & 50.9  & 54.9  & 66.1  & 48.5  & 46.3  & 59.7  & 71.5  & 51.4  & 48.6  & 53.9  & 39.9  & 44.1  & 51.9  \\ 
			Zhao et al.\cite{zhao2022graformer} CVPR' 22 & 45.2  & 50.8  & 48.0  & 50.0  & 54.9  & 65.0  & 48.2  & 47.1  & 60.2  & 70.0  & 51.6  & 48.7  & 54.1  & 39.7  & 43.1  & 51.8  \\ 
			Gong et al.\cite{gong2023diffpose} CVPR' 23 & \underline{42.8}  & \underline{49.1}  & \underline{45.2}  & \underline{48.7}  & 52.1  & 63.5  & \underline{46.3}  & \underline{45.2}  & 58.6  & \underline{66.3}  & 50.4  & 47.6  & 52.0  & {\bf 37.6}  & {\bf 40.2}  & 49.7  \\ 
			\midrule
			Ours (PGFormer) w/o aug & 45.4  & 51.5  & 46.1  & 50.0  & 51.1  & 59.4  & 48.5  & 46.5  & \underline{57.4}  & 67.6  & {\bf 49.4}  & 47.7  & 54.4  & 39.0  & 41.5  & 50.4  \\
			Ours (PGFormer) w/ aug & 43.8  & 49.9  & 45.5  & 49.0  & \underline{51.0}  & \underline{58.2}  & 47.0  & 45.7  & 58.1 & {\bf 65.8}  & \underline{49.6}  & {\bf 46.5}  & 53.6  & 38.0  & 40.6  & \underline{49.5} \\
			Ours (DiffPyramid) w/o aug & \textbf{42.3}  & \textbf{48.5}  & \textbf{44.4}  & \textbf{47.8}  & \textbf{50.8}  & 63.3  & \textbf{45.5}  & \textbf{44.8}  & \textbf{57.3}  & 66.6  & 49.8  & \underline{46.9}  & \underline{51.5}  & \underline{37.7}  & \underline{40.4}  & \textbf{49.2} \\
			\bottomrule
		\end{tabular}
	}
	\label{table:ResultSOTACPN}
\end{table*}

\begin{table*}[t]
	\caption{Results comparison with state-of-the-art frame-based methods on Human3.6M under MPJPE (mm). Ground truth 2D keypoints are used as input. The best and the second-best results are shown in bold and underlined, respectively. (+) uses extra data from MPII. (*) indicates the multi-scale methods. ({$\times$}) indicates the diffusion-based methods.}
	\resizebox{2\columnwidth}{!}{
		\begin{tabular}{*{17}{l}}
			\toprule
			Method
			&Dir.&Disc&Eat&Greet&Phone&Photo&Pose&Purch.&Sit&SitD.&Smoke&Wait&WalkD.&Walk&WalkT.
			&Avg ↓\\
			\midrule
			Martinez et al.\cite{martinez2017simple} ICCV'17 & 37.7  & 44.4  & 40.3  & 42.1  & 48.2  & 54.9  & 44.4  & 42.1  & 54.6  & 58.0  & 45.1  & 46.4  & 47.6  & 36.4  & 40.4  & 45.5  \\ 
			Zhao et al.\cite{zhao2019semantic} CVPR'19 & 37.8  & 49.4  & 37.6  & 40.9  & 45.1  & 41.4  & 40.1  & 48.3  & 50.1  & 42.2  & 53.5  & 44.3  & 40.5  & 47.3  & 39.0  & 43.8  \\ 
			Zhou et al.\cite{zhou2019hemlets} ICCV’19(+) &34.4 & 42.4 & 36.6 & 42.1 & 38.2 & 39.8 & 34.7 & 40.2 & 45.6 & 60.8 & 39.0 & 42.6 & 42.0 & 29.8 & 31.7 & 39.9\\
			Ci et al.\cite{ci2019optimizing} ICCV’19(+)(*)  & 36.3  & 38.8  & 29.7  & 37.8  & 34.6  & 42.5  & 39.8  & 32.5  & 36.2  &  \underline{39.5}  & 34.4  & 38.4  & 38.2  & 31.3  & 34.2  & 36.3  \\ 
			Liu et al.\cite{liu2020comprehensive} ECCV’20 & 36.8  & 40.3  & 33.0  & 36.3  & 37.5  & 45.0  & 39.7  & 34.9  & 40.3  & 47.7  & 37.4  & 38.5  & 38.6  & 29.6  & 32.0  & 37.8  \\ 
			Xu and Takano\cite{xu2021graph} CVPR'21(*) & 35.8  & 38.1  & 31.0  & 35.3  & 35.8  & 43.2  & 37.3  & 31.7  & 38.4  & 45.5  & 35.3  & 36.7  & 36.8  & 27.9  & 30.7  & 35.8  \\ 
			Zou and Tang\cite{zou2021modulated} ICCV'21 & 37.2  & 40.7  & 33.1  & 36.8  & 37.2  & 44.4  & 40.0  & 36.1  & 41.8  & 47.5  & 37.8  & 39.0  & 39.2  & 30.2  & 32.8  & 38.3  \\ 
			Zhao et al.\cite{zhao2022graformer} CVPR'22 & 32.0  & 38.0  & 30.4  & 34.4  & 34.7  & 43.3  & 35.2  & 31.4  & 38.0  & 46.2  & 34.2  & 35.7  & 36.1  & 27.4  & 30.6  & 35.2  \\ 
			Zhang et al.\cite{zhang2023learning} IJCV'23(*) & 32.4  & 36.5  & 30.1  & 33.3  & 36.3  & 43.5  & 36.1 & 30.5  & 37.5  & 45.3  & 33.8  & 35.1  & 35.3  & 27.5  & 30.2  & 34.9  \\ 
			Gong et al.\cite{gong2023diffpose}  CVPR' 23({$\times$}) & {\bf 28.8}  & {\bf 32.7}  & \underline{27.8} & \underline{30.9}  & \underline{32.8}  &  \underline{38.9} & {\bf 32.2} & \underline{28.3} & \underline{33.3} & 41.0 & \underline{31.0}  & {\bf 32.1}  &  \underline{31.5}  & \underline{25.9}  & \underline{27.5}  & \underline{31.6} \\
			\midrule
			Ours (PGFormer) & 32.3 & 36.2 & 30.5 & 32.6 & 33.8 & 42.4 & 36.3 & 29.1 & 36.0 & 43.9 & 33.9 & 34.5 & 35.1 & 26.6 & 29.6 & 34.2 \\ 
			Ours (DiffPyramid) ({$\times$}) & \underline{29.2} & \underline{33.5} & {\bf 26.1} & {\bf 29.9} & {\bf 32.2} & {\bf 35.7} & \underline{32.9} & {\bf 28.0} & {\bf 31.6} & {\bf 38.9} & {\bf 30.4} & \underline{32.2} & {\bf 31.4} & {\bf 25.8} & {\bf 26.7} & {\bf 31.0} \\
			\bottomrule
		\end{tabular}	}
	\label{table:ResultSOTAGT}
\end{table*}
\subsection{Datasets and Evaluation Metrics}
We evaluate our approach on two standard 3D HPE benchmarks: Human3.6M \cite{ionescu2013human3} and MPI-INF-3DHP \cite{mehta2017monocular}.\par
{\it Human3.6M} is the most widely used dataset for 3D HPE task, containing 3.6 million images and 15 daily activities. There were 11 human subjects filmed in an indoor environment by 4 synchronized cameras in different views. In the past, there were two common evaluation protocols used.
Protocol \#1 uses Mean Per Joint Position Error (MPJPE) in millimeter as evaluation metric, which evaluates the average Euclidean distance between the predicted joint and the ground truth in millimeters after aligning the root joint (the hip joint). Protocol \#2 is calculated after rigid transformation.
Following \cite{zhao2019semantic,zou2021modulated,liu2020comprehensive,martinez2017simple,xu2021graph}, we use five subjects (S1, S5, S6, S7 and S8) for training and two subjects (S9 and S11) for testing under Protocol \#1. Because it is more challenging and Protocol \#2 shows the same trend as Protocol \#1.\par
{\it MPI-INF-3DHP} is another challenging 3D pose dataset with both indoor and outdoor scenes. Its rich action and scenes complement the shortcomings of Human 3.6M. We evaluate the generalization on its testset. {Following \cite{ci2019optimizing,zhou2019hemlets,zou2021modulated,xu2021graph}, 3D percentage of correct keypoints (3DPCK) under the threshold of 150 mm, and area under curve (AUC) are adopted as metrics.}
\subsection{Implementation details}
{In order to process multi-view Human3.6M data, we followed the method of \cite{zhao2019semantic,zhao2022graformer} for normalization processing. To tackle uncertainties in 2D detector data, we set separate hyperparameters for ground truth and detected data for better convergence. For 2D detector data, we use the cascaded pyramid network (CPN) \cite{lin2017FPNobject} for 2D pose detection.  We set the hidden layer size to 256 with 4 layers as shown in Fig. \ref{fig:network architecture}. The learning rate starts at 0.001 and decays by 4\% every 4 epochs. The loss $\lambda$ is set to 0.1. Our model is trained for 15 epochs with batch size 256. Data augmentation includes only horizontal flip \cite{cai2019exploiting,zou2021modulated,li2022mhformer}. For 2D ground truth data, the hidden layer size is set to 96 with 5 layers. The learning rate starts at 0.001 and is multiplied by 0.9 every 50,000 steps. Our model is trained for 50 epochs with mini-batches of 64. The  $\lambda$ is set to 0.025.}\par
To train the proposed DiffPyramid, we set the batchsize to 2048. {Human3.6M dataset is formatted into GMM according to the data processing in \cite{gong2023diffpose}. The diffusion inference for all experiments is accelerated using DDIM \cite{ddim}, completing reverse diffusion in five steps. $\lambda$ is set to the same as PGFormer. The initial learning rate is 0.0005, decaying by 0.9 after 10 epochs. In the forward diffusion process, the $\beta_{1:K}$ is from $1e-4$ to $2e-3$ via a linear function, the sample number N  is 5 and the number of reverse diffusion steps K to 50. DiffPyramid does not use horizontal flipping.
Other hyperparameters and training details match those of PGFormer.}
\subsection{Comparison with State-of-the-Art Methods}
{\it Results on Human3.6M.} We compare our model with
some state-of-the-art methods on Human3.6M under MPJPE. Our methods are compared with the advanced single-frame methods under CPN as Table \ref{table:ResultSOTACPN}. Some methods use multiple video frames as input or apply pose refinement in post-processing. Note that the improvement of PGFormer compared with baseline GraFormer \cite{zhao2022graformer} is significant. The error was reduced from 51.8 to 49.5 (relative 4\%). And it can be seen that our method is also comparable to some temporal methods. { So the proposed lightweight and fast PGFormer can be used effectively in applications. We observe that PGFormer achieves strong performance without data augmentation, with optimal results obtained by learning the symmetric prior of the human body through horizontal flipping. And DiffPyramid, a variant of the diffusion model,  achieves better accuracy and outperforms all single-frame methods.}\par 
In addition, we report the results in Table \ref{table:ResultSOTAGT} for five aspects. (1) Among methods within the same category, the DiffPyramid is the best when using the 2D keypoints ground truth as input. (2) Compared with methods considering {\it long-range dependence}, such as MGCN \cite{zou2021modulated}, PGFormer exhibits an error reduction of {4.1 mm}. It corresponds to a relative improvement of {10.7\%}, indicating better modeling of long-range dependence. And PGFormer is good for actions with long range dependence, such as Photo, Walk and SittingDown (adjacent nodes are farther away than other actions).  (3) PGFormer achieves the optimal result of {34.2 mm}, surpassing other {\it multi-scale} methods such as GraphSH \cite{xu2021graph} (35.8 mm) and PHGANet \cite{zhang2023learning} (34.9 mm). (4) We observe that multi-scale methods exhibit relatively lower errors in {\it complex actions}. 
For example, our proposed PGFormer demonstrates a notable enhancement, achieving {33.9} mm on 'Smoke' (vs 53.5 mm by SemGCN \cite{zhao2019semantic} and 37.8 mm by MGCN) and 36.0 mm on 'Sit' (vs 50.1 mm by SemGCN and 41.8 mm by MGCN). In other words, human sub-structure becomes more pronounced in complex actions. (5) We acknowledge that PGFormer does not perform as well as the current diffusion model \cite{gong2023diffpose}. But a variation of its diffusion model, i.e. DiffPyramid, achieves better precision, which also explains the versatility of PGFormer. It also shows that the proposed hierarchical long-range dependencies is also required by the diffusion model. 
\par 
In summary, it means that PGFormer has a stronger ability to capture multi-scale long-range spatial constraints to reduce deep ambiguity and errors in 3D HPE. Moreover, the advantages mentioned above are consistent for the DiffPyramid, which also explains the generality of our model.\par
{\it Comprehensive Performance.} In order to evaluate the effectiveness, we consider comparing both parameters and performance to measure the comprehensive performance of the models, shown in Table \ref{table:ResultParams}. PGFormer achieves better results with even much fewer parameters than Pre-agg, GraphSH and PHGANet. PGFormer utilizes only 0.66M parameters, resulting in a relative reduction of 84.3\%, 82.1\%, and 77.5\%, respectively. {Additionally, the version of DiffPyramid we provide offers improved accuracy with a certain sacrifice in the number of parameters. These two versions demonstrate superior performance and resource efficiency, enabling easier deployment and scalability in practical applications.}\par
\begin{figure}[t]
\begin{minipage}[b]{1.0\linewidth}
  \centering
  \centerline{\includegraphics[width=9cm]{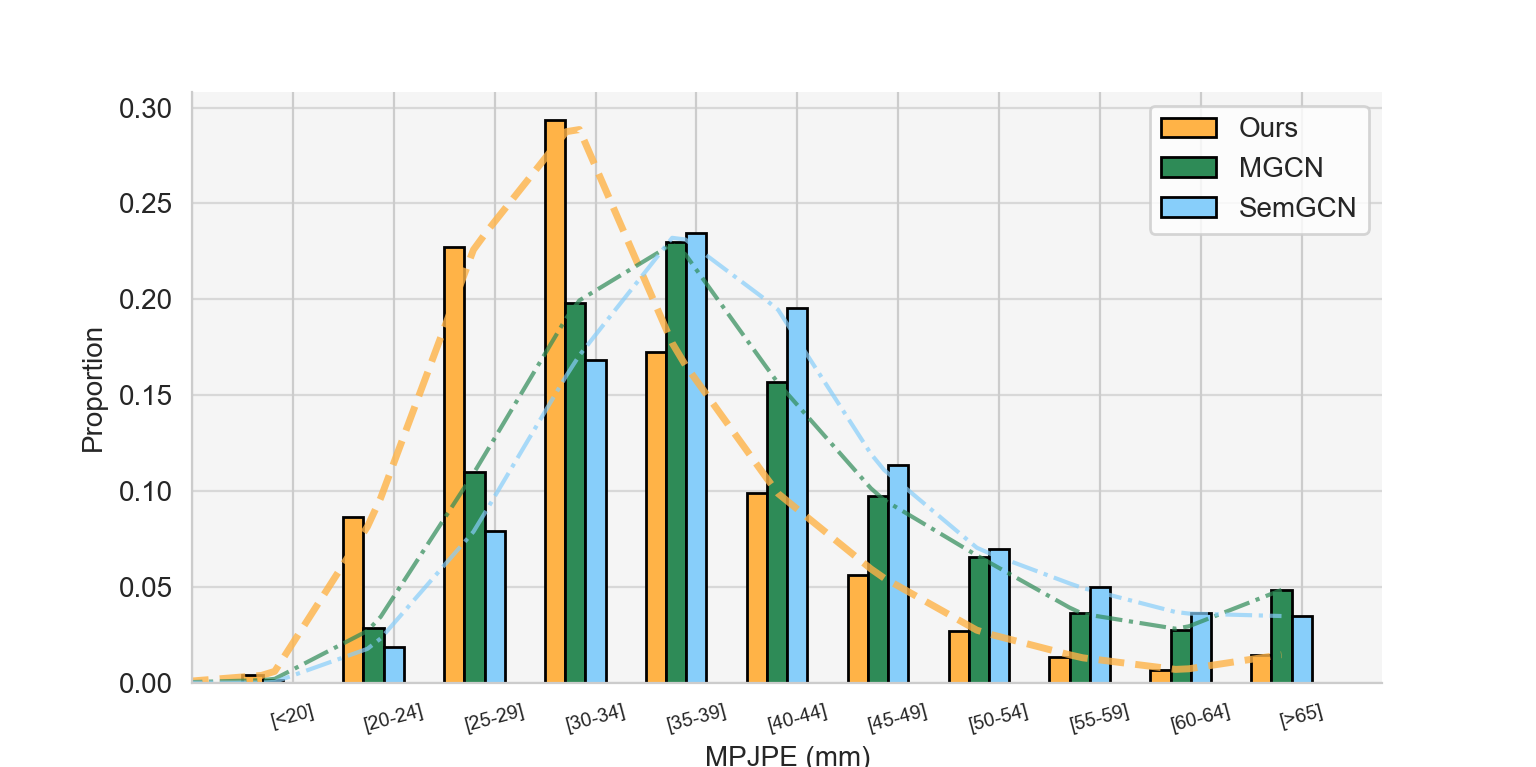}}
  \medskip
\end{minipage}
\caption{MPJPE distribution on the testset of Human3.6M. We compare PGFormer with the other two methods. Our distribution indicates enhanced robustness and superiority of the method.}
\label{fig:Distribution}
\end{figure}

\begin{table}[htb]
\caption{Results on Human3.6M dataset with parameter and performance. (*) indicates the multi-scale methods. Our method has more advantages in comprehensive performance, especially compared with the existing multi-scale methods.}
\resizebox{1\columnwidth}{!}{
\begin{tabular}{*{3}{c}}
\toprule
Method &  params ↓ & MPJPE(mm) ↓ \\
\midrule
GAT(Velickovic et al. 2017\cite{velickovic2017graph}) & 0.16M & 82.9 \\ 
FC(Martinez et al. 2017\cite{martinez2017simple}) & 4.29M & 45.5 \\ 
SemGCN(Zhao et al. 2019\cite{zhao2019semantic}) & 0.43M & 40.8 \\ 
Pre-agg(Liu et al. 2020\cite{liu2020comprehensive}) & 4.20M & 37.8 \\  
MGCN(Zou and Tang 2021\cite{zou2021modulated}) & 0.29M & 38.3 \\ 
GraFormer(Zhao et al.2022\cite{zhao2022graformer}) & 0.65M & 35.2 \\ 
\midrule
GraphSH(Xu and Takano 2021\cite{xu2021graph})(*) & 3.70M & 35.8 \\
PHGANet(Zhang et al.2023\cite{zhang2023learning})(*) & 2.94M & 34.9 \\
\midrule
PGFormer (Ours) & 0.66M & 34.2 \\
DiffPyramid (Ours) & 1.86M & 31.0 \\
\bottomrule
\end{tabular}
}
\label{table:ResultParams}
\end{table}
{\it Distribution of MPJPE.} We further explore the robustness of PGFormer. The distribution in Fig. \ref{fig:Distribution} shows that 
PGFormer exhibits a smaller mean and variance of errors. Especially, a higher proportion of errors is below 35 mm, while errors exceeding 50 mm are less frequent. It means that PGFormer has better robustness and performance on complex actions. It is able to maintain sufficient stability without estimating excessively large errors. This also shows that the performance improvement of PGFormer is comprehensive, which not only reduces the overall mean error, but also reduces the variance.\par
{\it Results on MPI-INF-3DHP.} We train PGFormer on the Human3.6M and test it directly on the MPI-INF-3DHP to assess the generalization of PGFormer. The results are reported in Table \ref{table:ResultSOTA3DHP}. Following \cite{gong2021poseaug,li2022mhformer}, we used ground truth on Human3.6M training results when conducting cross-dataset generalization tests. Our method outperforms existing single-frame approaches in both indoor and outdoor environments, highlighting its improved generalization.\par
\begin{table}[htb]
\caption{Quantitative comparison of the MPI-INF-3DHP dataset. }
\resizebox{1\columnwidth}{!}{
\begin{tabular}{*{7}{c}}
\toprule
Method & Training data & 3DPCK(GS) & no GS & Outdoor & All (PCK) ↑ & All (AUC) ↑\\
\midrule
Zhou et al.\cite{zhou2017towards} & H36M+MPI & 71.1 & 64.7 & 72.7 & 69.2 & 32.5 \\ 
Yang et al.\cite{yang20183d} & H36M+MPI & – & – & – & 69 & 32 \\ 
Zhou et al.\cite{zhou2019hemlets} & H36M & 75.6 & 71.3 & 80.3 & 75.3 & 38 \\ 
Pavllo et al.\cite{pavllo:videopose3d:2019} & H36M & 76.5 & 63.1 & 77.5 & 71.9 & 35.3 \\ 
Ci et al.\cite{ci2019optimizing} & H36M & 74.8 & 70.8 & 77.3 & 74 & 36.7 \\ 
Liu et al.\cite{liu2020comprehensive} & H36M & 77.6 & 80.5 & {\bf 80.1} & 79.3 & 47.6 \\ 
Xu and Takano\cite{xu2021graph} & H36M & 81.5 & 81.7 & 75.2 & 80.1 & 45.8 \\ 
Zhao  et al.\cite{zhao2022graformer} & H36M & 80.1 & 77.9 & 74.1 & 79 & 43.8 \\ \hline
Ours w/o aug & H36M & 83.2 & 83.5 & 76.8 & 82.4 & 51.4\\ 
Ours w/ aug & H36M & {\bf 84.4} & {\bf 84.5} & 77.4 & {\bf 83.9} & {\bf 52.3} \\ 
\bottomrule
\end{tabular}
}
\label{table:ResultSOTA3DHP}
\end{table}
{\it Inference Runtime Results.} We tested the performance of our method on real-time systems using a standard desktop computer equipped with an Intel(R) Core(TM) i7-9750H CPU @ 2.60GHz, 16 GB of RAM, and a GTX 1050 graphics card. The inference runtime, shown in Table \ref{table:Real-TimePerformance}, was evaluated with a batch size of 64 and 12 workers, using hyperparameter settings consistent with the original paper, with no horizontal flip applied during testing. PGFormer completes inference in just 0.02 seconds per batch of 64, achieving a speed of 3200 fps, which is 10 times faster than the diffusion model method \cite{gong2023diffpose} (3200 fps vs. 98 fps). Moreover, the accuracy improvement does not compromise speed compared to the baseline \cite{zhao2022graformer}. Our method also significantly outperforms time-series approaches \cite{li2022mhformer} in speed, fully meeting real-time requirements. Additionally, tests on both CPU and GPU demonstrate that our approach achieves real-time performance across different devices. {Meanwhile, our proposed DiffPyramid version meets basic real-time requirements, capable of processing approximately 95 frames per second on the GPU. In summary, our method meets most real-time requirements and allows for the selection of an appropriate version based on the desired trade-off between accuracy and speed.}\par
\begin{table}[htb]
	\centering
	\caption{Quantitative Real-Time Performance on Human3.6M Dataset using a standard desktop. The size of One Batch is set to 64. The experiment shows the advantage of the inference time of our method. (†) uses temporal information.}
	\resizebox{0.9\columnwidth}{!}{
		\begin{tabular}{*{7}{c}}
			\toprule
			Method & Device & Time Average(One Batch)↓\\
			\midrule
			Li et al.\cite{li2022mhformer} (†) & GPU & 3290ms\\
			Zhao et al.\cite{zhao2022graformer} & GPU & 15ms\\
			Gong et al.\cite{gong2023diffpose} & GPU & 652ms\\
			\midrule
			Ours (PGFormer) & CPU & 68ms\\ 
			Ours (PGFormer) & GPU & 19ms\\ 
			Ours (DiffPyramid) & GPU & 661ms\\
			\bottomrule
		\end{tabular}
	}
	\label{table:Real-TimePerformance}
\end{table}
\subsection{Ablation Study}
{We conduct additional ablation experiments on Human3.6M under MPJPE. We use ground truth as input data to prevent the influence of error caused by 2D pose detection. This fully illustrates the improved performance of our method and the role of each ablation.}\par
{\it Components of Pyramid Graph Attention.} To explore the effectiveness of components in our model, we employ GraFormer \cite{zhao2022graformer} as the baseline method, which does not consider multi-scales. The overall ablation experiment of the Pyramid Graph Attention module is presented in Table \ref{table:AblationFusionPGAmodule}.\par 
{First, we designed a pure spatial transformer with multi-head attention and MLP layers, but its performance was poor due to lacking a skeletal prior. The Pyramid Transformer, with PGA modules, improved this by capturing skeletal priors effectively. However, global attention and local convolution alone were not efficient enough. Combining both approaches led to superior results, highlighting that local priors and long-range dependencies are crucial to understanding the human body's prior knowledge. Additionally, the integration of pyramid attention and graph convolution enhances the model's ability to capture prior constraints and reduce errors.}\par
\begin{table}[htb]
	\tiny 
	\renewcommand\arraystretch{0.5}
	\centering
	\caption{Ablation study of Pyramid Graph Attention module}
	\label{table:AblationFusionPGAmodule}
	\resizebox{1\columnwidth}{!}{
		\begin{tabular}{*{7}{c}}
			\toprule
			ChebGCN \cite {defferrard2016convolutional} & Transformer & Pyramid Transformer & MPJPE (mm) \\
			\midrule
			$\times$ & \checkmark & $\times$ & 51.7 \\
			\rowcolor{mygray} \checkmark & $\times$ & $\times$ & 47.8\\
			$\times$ & $\times$ & \checkmark & 41.8\\
			\rowcolor{mygray} \checkmark & \checkmark & $\times$ & 35.2 \\
			\checkmark & $\times$ & \checkmark & 34.5 \\
			\bottomrule
		\end{tabular}
	}
\end{table}
In Table \ref{table:AblationFusionPoolingScale} and \ref{table:AblationFusionNumberPooling}, we explore the different parts of pooling in Eq. \ref{equation:Pooling} and Eq. \ref{equation:PoolingConcat} to verify the validity of the pyramid structure in our method. First,
we explored adaptive pooling methods\cite{pmlr-v97-lee19c,ma2020path} and varying pool sizes at different scales for grouped pooling within a pyramid structure in Table \ref{table:AblationFusionPoolingScale}. We found that adaptive pooling struggles to learn effective grouping features. While these methods work well for general graph data, the skeleton graph's unique structure makes adaptive pooling less effective than manual grouping. Beacause the information is transmitted through bones and explicit node connections In this experiment, only a single layer of pooled features was fused.\par
\begin{figure}[htb]
	\begin{minipage}[b]{1\linewidth}
		\centering
		\centerline{\includegraphics[width=8cm]{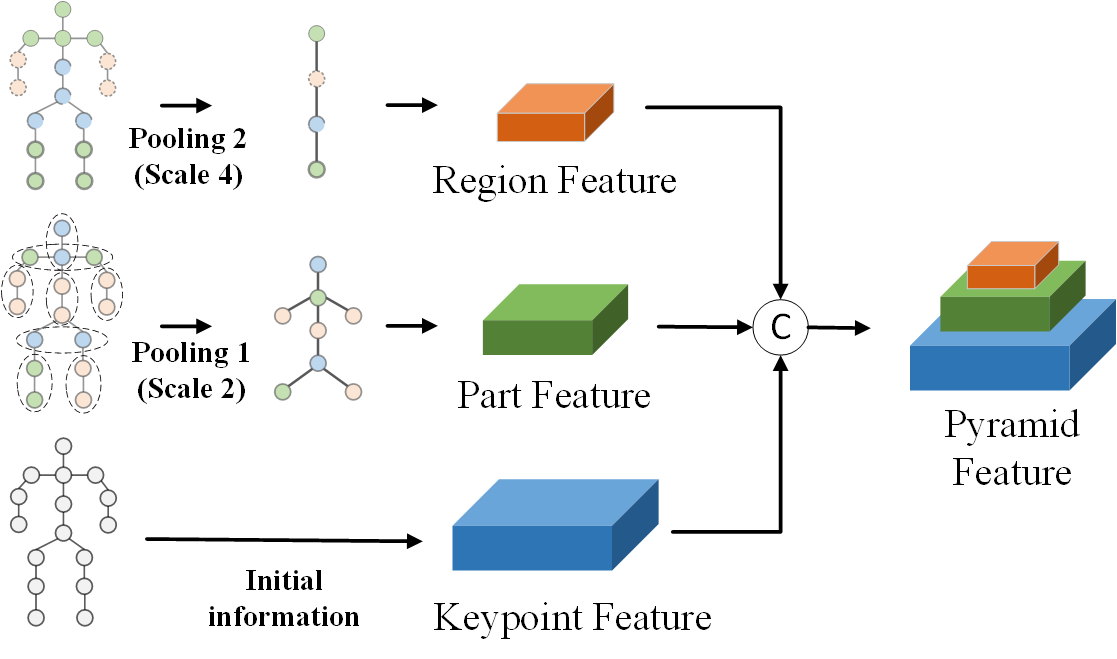}}
		\medskip
	\end{minipage}
	\caption{A diagram illustrating the pooling strategy and construction of pyramid structure.}
	\label{fig:pyramidPoolingExp}
\end{figure}
{In particular, we designed the pooling layer, as shown in Fig. \ref{fig:pyramidPoolingExp}, based on \cite{xu2021graph, zhao2019semantic}, which is a concrete implementation of Eq. \ref{equation:Pooling}. Each manually defined pool size corresponds to skeletal connections in the human body, segmenting it into distinct body parts. Specifically, Pooling 1 reduces the 16 joints to 8, segmenting the body into two legs, arms, head, torso, and hips with a scale factor of two. Pooling 2, with a scale factor of four, further reduces the keypoints to four regions: head, hands, torso, and legs.}\par

\begin{figure}[htb]
	\begin{minipage}[b]{1\linewidth}
		\centering
		\centerline{\includegraphics[width=6cm]{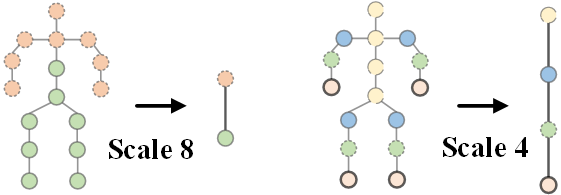}}
		\medskip
	\end{minipage}
	\caption{A diagram displaying the other scales of pooling.}
	\label{fig:otherpooling}
\end{figure}

As shown in Table \ref{table:AblationFusionPoolingScale}, manual grouping helps the model better learn the human body's structure, though the impact of a single pooling layer is relatively small. Specifically, the grouping at scale 8, shown in Fig. \ref{fig:otherpooling}, has a noticeable effect. For scale 4, we tested two grouping strategies (see Fig. \ref{fig:pyramidPoolingExp} and Fig. \ref{fig:otherpooling}), yielding similar results. Repeating the experiment, the error ranged between 35.4 and 35.0. Pooling at scale 2 further improved the model, reducing the error from 35.2 to 34.9. Thus, a well-chosen pooling layer enhances the model’s performance through the PGA module.

\begin{table}[htb]
	\tiny 
	\renewcommand\arraystretch{0.5}
	\centering
	\caption{Ablation study for pooling scale}
	\label{table:AblationFusionPoolingScale}
	\resizebox{1\columnwidth}{!}{
		\begin{tabular}{*{7}{c}}
			\toprule
			Method & MPJPE (mm) & Method & MPJPE (mm) \\
			\midrule
			w/o Pooling & 35.2 & Pooling scale 2 & 34.9\\
			\rowcolor{mygray} SAGPool\cite{pmlr-v97-lee19c} & 36.9 & Pooling scale 4 & 35.0 \\
			PanPool\cite{ma2020path} & 36.5 & Pooling scale 8 & 35.6 \\
			\bottomrule
		\end{tabular}
	}
\end{table}

In Tables \ref{table:AblationFusionNumberPooling} and \ref{table:AblationFusionPyramidWay}, we construct the feature pyramid using pooling layers and evaluate the impact of varying the number of layers and feature fusion methods. First, we designed the pyramid structure shown in Fig. \ref{fig:pyramidPoolingExp}. As shown in Table \ref{table:AblationFusionNumberPooling}, learning pyramid features reduced the error from 35.20 to 34.54, improving accuracy by 1.91\%. The increase in model size due to our method is minimal. These results demonstrate that pooling across different scales enhances the learning of human sub-structural information, significantly boosting the effectiveness of self-attention in the model. Overall, constructing a feature pyramid yields a substantial performance improvement.\par

We further explore the impact of various pyramid feature fusion methods in Table \ref{table:AblationFusionPyramidWay}. In this experiment, we kept the pooling scale at 2 and added only one pooling layer for ablation. For the fusion experiments, which involved summation and multiplication, we applied pooling followed by linear interpolation upsampling to maintain consistent scale sizes during computation. The fusion method of pooling followed by unpooling led to information loss and struggled to capture effective features. In contrast, concatenating channels from different scales yielded the best performance, as it preserved information without any loss.\par

\begin{table}[htb]
	\begin{minipage}[b]{0.48\linewidth}
		\tiny 
		\renewcommand\arraystretch{0.5}
		\centering
		\caption{Ablation study for number of pooling}
		\label{table:AblationFusionNumberPooling}
		\resizebox{1\columnwidth}{!}{
			\begin{tabular}{*{7}{c}}
				\toprule
				Method & MPJPE (mm) \\
				\midrule
				w/o Pooling & 35.2 \\
				Pooling 1 & 34.9 \\
				Pooling 1\&2 & 34.5 \\
				\bottomrule
			\end{tabular}
		}
	\end{minipage}
	\begin{minipage}[b]{0.48\linewidth}
	\tiny 
	\renewcommand\arraystretch{0.5}
	\centering
	\caption{Ablation study on different ways of feature fusion}
	\label{table:AblationFusionPyramidWay}
	\resizebox{1\columnwidth}{!}{
		\begin{tabular}{*{7}{c}}
			\toprule
			Method & MPJPE (mm) \\
			\midrule
			Pyramid Add & 36.4\\
			Pyramid Mul & 36.8\\
			Pyramid Concat & 34.9\\
			\bottomrule
		\end{tabular}
	}
	\end{minipage}
\end{table}

{\it Pyramid Graph Attention module.} To further verify the universality of the Pyramid Graph Attenti on (PGA) module, we designed a plug-and-play ablation experiment of the module. We respectively use SemGCN \cite{zhao2019semantic}, Modulated GCN \cite{zou2021modulated} and ChebGCN \cite{defferrard2016convolutional} as graph convolution layers. The outcomes presented in Table \ref{table:AblationDGCN} demonstrate the universality of our approach. It demonstrates that our versatile and flexible method can serve as a plug-and-play module for integrating various graph convolution methods.\par 
\begin{table}[htb]
	\tiny
	\renewcommand\arraystretch{0.5}
	\centering
	\caption{Ablation study on various graph convolution methods}
	\label{table:AblationDGCN}
	\resizebox{0.9\columnwidth}{!}{
		\begin{tabular}{lcc}
			\toprule
			GCN Method & Params & MPJPE (mm) \\
			\midrule
			SemGCN \cite{zhao2019semantic} & 0.27M & 52.5 \\ 
			\rowcolor{mygray}\ \ \ \ \ \  +Ours & 0.20M & 38.2 \\
			Modulated GCN \cite{zou2021modulated} & 0.29M & 38.3 \\
			\rowcolor{mygray}\ \ \ \ \ \ +Ours & 0.22M & 36.8 \\
			ChebGCN \cite{defferrard2016convolutional} & 0.65M & 35.2 \\
			\rowcolor{mygray}\ \ \ \ \ \ +Ours & 0.66M & 34.5 \\
			\bottomrule
		\end{tabular}
	}
\end{table}
GCN methods are limited by the skeletal graph’s natural connections, whereas our method complements long-range dependencies in human body structure. This is most effectively demonstrated by the experiment with SemGCN, which aggregates only first-order features and completely ignores long-range dependencies. After integrating the PGA module, SemGCN learns long-range dependency information effectively, reducing the error from 52.5 to 38.2 (relative reduction of 27\%). Additionally, GCN methods that aggregate higher-order information also show significant performance improvements after adding the PGA module. For example, MGCN's error is reduced by 3.9\%. The results indicate that ChebGCN and Modulated GCN have varying abilities to aggregate long-range, higher-order nodes. While the improvement from modeling long-distance dependencies is less pronounced for these methods, it still represents a substantial enhancement.\par
It is important to note that the Pyramid Graph Attention and non-local \cite{wang2018non} modules both aim to capture long-range dependencies and are not compatible. Therefore, we used SemGCN without the non-local module for a fair comparison. The dimensions of SemGCN’s hidden layers were adjusted to match the GraphSH experiment. In contrast, MGCN was configured with the non-local module, as in the original paper. This setup effectively demonstrates the effectiveness of the proposed Pyramid Graph Attention module.\par

\begin{table}[htb]
	\footnotesize
	\centering
	\caption{Comparison with other multi-scale methods}
	\resizebox{1\columnwidth}{!}{
		\begin{tabular}{cccc}
			\toprule
			GCN Method & Frameworks & Params & MPJPE (mm) \\ 
			\midrule
			SemGCN & SeqRes & 0.43M & 40.7 \\ 
			\rowcolor{mygray} ~ & GraphSH\cite{xu2021graph} & 0.44M & 39.2 \\ 
			~ & Parallel multi-scale\cite{zhang2023learning} & 0.79M & 38.7 \\ 
			\rowcolor{mygray} ~ & PGFormer (Ours) & 0.20M & 38.2 \\
			\midrule
			Modulated GCN & SeqRes & 0.29M & 38.3 \\ 
			\rowcolor{mygray} ~ & GraphSH & - & - \\ 
			~ & Parallel multi-scale & 2.94M & 36.4 \\ 
			\rowcolor{mygray} ~ &  PGFormer (Ours) & 0.22M & 36.8 \\ 
			\bottomrule
		\end{tabular}
	}
	\label{table:ComparisonDiffGCN}
\end{table}
Meanwhile, the results in Table \ref{table:ComparisonDiffGCN} show that our method outperforms other multi-scale approaches, demonstrating that pose prediction benefits from incorporating multi-scale semantic features in a parallel manner. Specifically, we achieve similar accuracy to PHGANet \cite{zhang2023learning} while using only about 22\% of the parameters to learn multi-scale features. This indicates that learning multi-scale features at different depths or branches is not as efficient. Instead, using the self-attention mechanism to capture cross-scale correlations is sufficient to enhance the modeling of human substructures.\par
\begin{table}[htb]
	\footnotesize
	\centering
	\caption{Comparison with other multi-scale methods}
	\resizebox{0.9\columnwidth}{!}{
		\begin{tabular}{cccc}
			\toprule
			Spatial condition & Diffusion model & MPJPE (mm) \\ 
			\midrule
			GraFormer\cite{zhao2022graformer} & - & 35.6 \\
			\rowcolor{mygray} &  DiffPose\cite{gong2023diffpose} & 31.6 \\
			& DiffPyramid (Ours) & 31.2 \\
			\midrule
			\rowcolor{mygray} Ours (PGFormer) & - & 34.7\\
			& DiffPose & 31.5\\
			\rowcolor{mygray} & DiffPyramid (Ours) & 31.0\\
			\bottomrule
		\end{tabular}
	}
	\label{table:DiffPyramid}
\end{table}
To verify the effectiveness of the proposed method within the diffusion model architecture, we conducted modular ablation experiments using the designed DiffPyramid, as illustrated in Table \ref{table:DiffPyramid}. First, it is difficult to predict 3D pose without a priori spatial conditions, which results in a very large error (e.g. 2713 mm MPJPE). In Table \ref{table:DiffPyramid}, it is evident that the PGA module within DiffPyramid effectively extracts features from various human substructures, thereby facilitating the learning of spatial distributions. Ultimately, the strongest results are obtained through the more robust priors provided by PGFormer.

{\it Regularization.} We perform l1-norm ablation experiments by using the values of ablation $\lambda$. We conducted experiments on CPN data and ground truth data respectively. We found that by adding l1-norm, the sparsity of the model can be increased, making the learned matrix closer to 0, which further strengthens the extraction of nodes that play an important role in long-distance dependencies. This improves the learning efficiency of the model and mitigates overfitting.\par

\begin{table}[htb]
	\footnotesize
	\centering
	\caption{Ablation study for $l1-norm$.}
	\resizebox{1\columnwidth}{!}{
		\begin{tabular}{*{7}{c}}
			\toprule
			Method on CPN & MPJPE & Method on GT & MPJPE  \\
			\midrule
			w/o PGA module & 51.8 & w/o PGA module & 35.2\\
			PGA + w/o $l1-norm$ & 49.9  & PGA + w/o $l1-norm$ & 34.5 \\
			\midrule
			$\lambda$=0.025 & 49.9 & {\bf $\lambda$=0.025} & {\bf 34.2} \\
			$\lambda$=0.05 & 49.7  & $\lambda$=0.05 & 34.7 \\
			{\bf $\lambda$=0.1} & {\bf 49.5} & $\lambda$=0.1 & 35.0 \\
			$\lambda$=0.2 & 50.4 & $\lambda$=0.2 & 35.6 \\
			\bottomrule
		\end{tabular}
	}
	\label{table:AblationL1}
\end{table}

\subsection{Visualization}
\label{sec:typestyle}
\begin{figure}[htb]
	\centering
	\centerline{\includegraphics[width=0.5\textwidth]{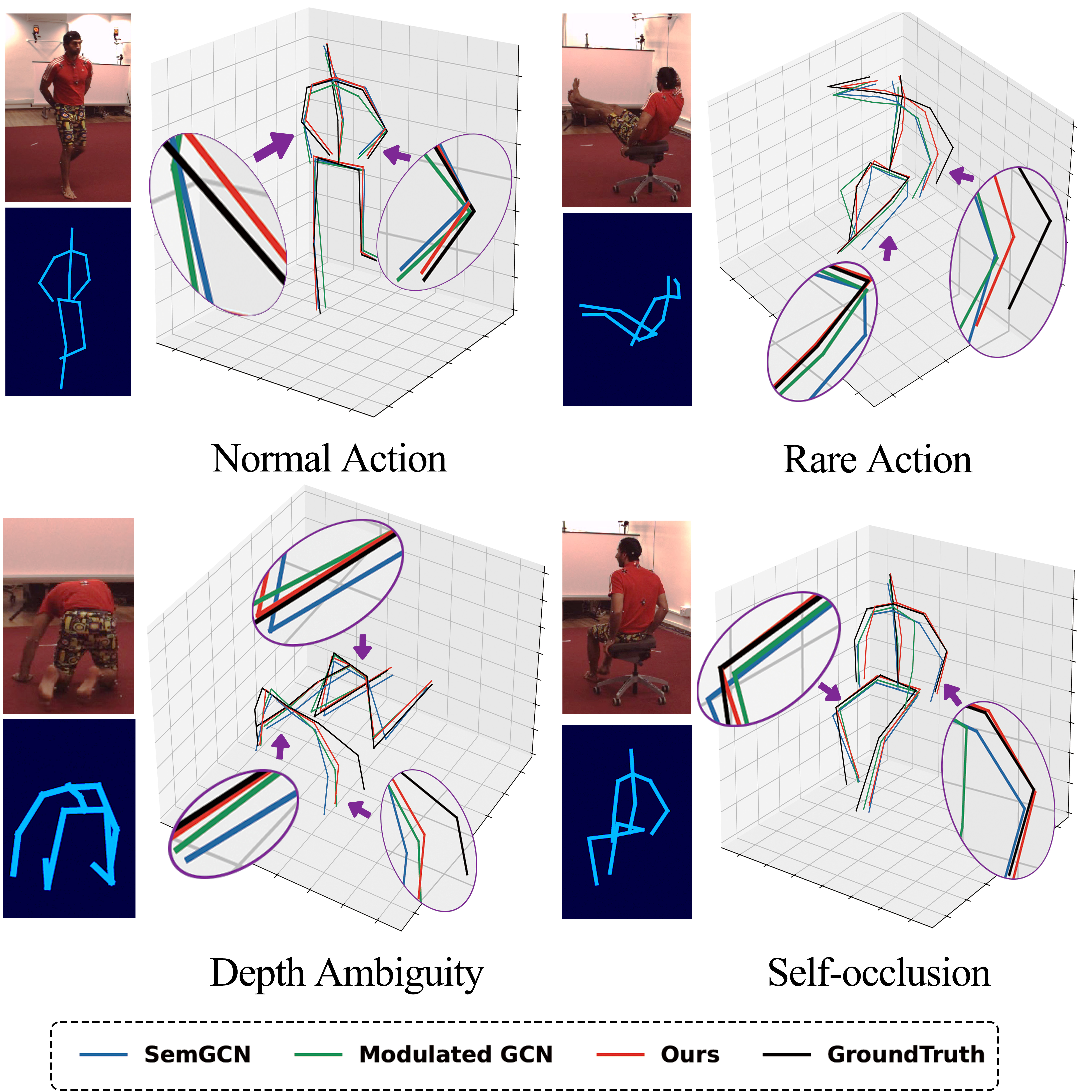}}
	\medskip
	\caption{Qualitative comparisons on Human3.6M dataset. Ground truth is the black line. The result of SemGCN is the blue line. The result of MGCN is the green line. PGFormer is shown in the red line. Our approach is closer to groundtruth, proving its superior performance benefits. It is also robust and reliable in complex actions.}
	\label{fig:H36MResultShow}
\end{figure}

To qualitatively evaluate our method, Fig. \ref{fig:H36MResultShow} provides a visual comparisons on Human3.6M between PGFormer and two state-of-the-art approach, SemGCN\cite{zhao2019semantic} and MGCN\cite{zou2021modulated}. It is obvious that our model achieves excellent performance and robustness.\par
We visualize walking and sitting actions in Human3.6M. The results show that PGFormer can provide reasonable 3D prediction results on both normal actions and the actions with deep ambiguity and self-occlusion. The visualized results align with our ablation and comparison experiments. It means that PGFormer exhibits more reliable results and demonstrates robust performance in complex actions. Further analysis, SemGCN is a method that considers the adjacent keypoints, while long-range dependencies in MGCN are mined through a learning matrix. And our method fully models the long-range dependencies. This long-range dependence is especially reflected in the hands and feet, which are far from the center of the hip. The remote keypoints are closer to the exact result under the constraint of long-range relations.\par

\begin{figure}[t]
	\centering
	\centerline{\includegraphics[width=0.5\textwidth]{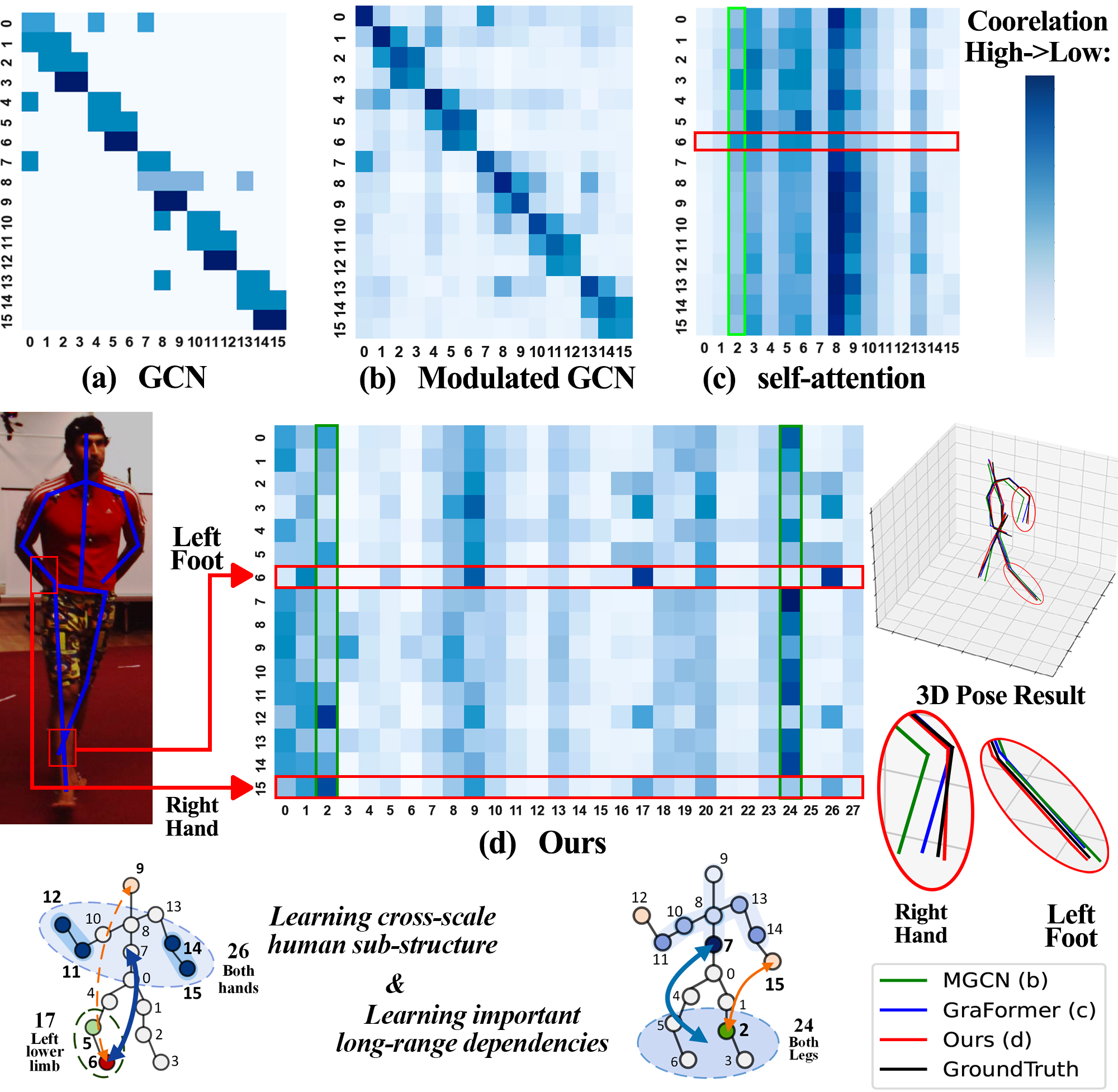}}
	\medskip
	\caption{Visualization of Pyramid Graph Attention on Human3.6M dataset. The proposed method effectively captures cross-scale long-range dependencies and human sub-structures, learning distinctive and important correlations. For example, the correlation between right hand (Joint 15) and left foot (Joints 2) and the effect of the lower body (Joint 24) on the upper body joints.}
	\label{fig:AttentionMap}
\end{figure}

{In addition, we visualize in Fig. \ref{fig:AttentionMap} whether PGFormer has learned effective human structural features instead of noise introduced by coordinates.
	Fig. \ref{fig:AttentionMap} (a) displays the normalized graph convolution adjacency matrix, highlighting features constrained by the physical structure of the human body.
	Fig. \ref{fig:AttentionMap} (b) shows the long-range dependencies learned from Modulated GCN \cite{zou2021modulated}, which can only capture minor long-distance dependencies, rather than critical information.
	Fig. \ref{fig:AttentionMap} (c) illustrates that the standard self-attention mechanism ignores human structure and introduces noise based on coordinate values.
	Fig. \ref{fig:AttentionMap} (d) demonstrates the ability of Pyramid Graph Attention to learn human structure and long-range dependencies, including sub-structural features and constraints beyond the natural skeleton graph. It can be observed that the learned features are more efficient and critical.}
\par

\begin{figure}[htb]
	\centering
	\includegraphics[width=0.5\textwidth]{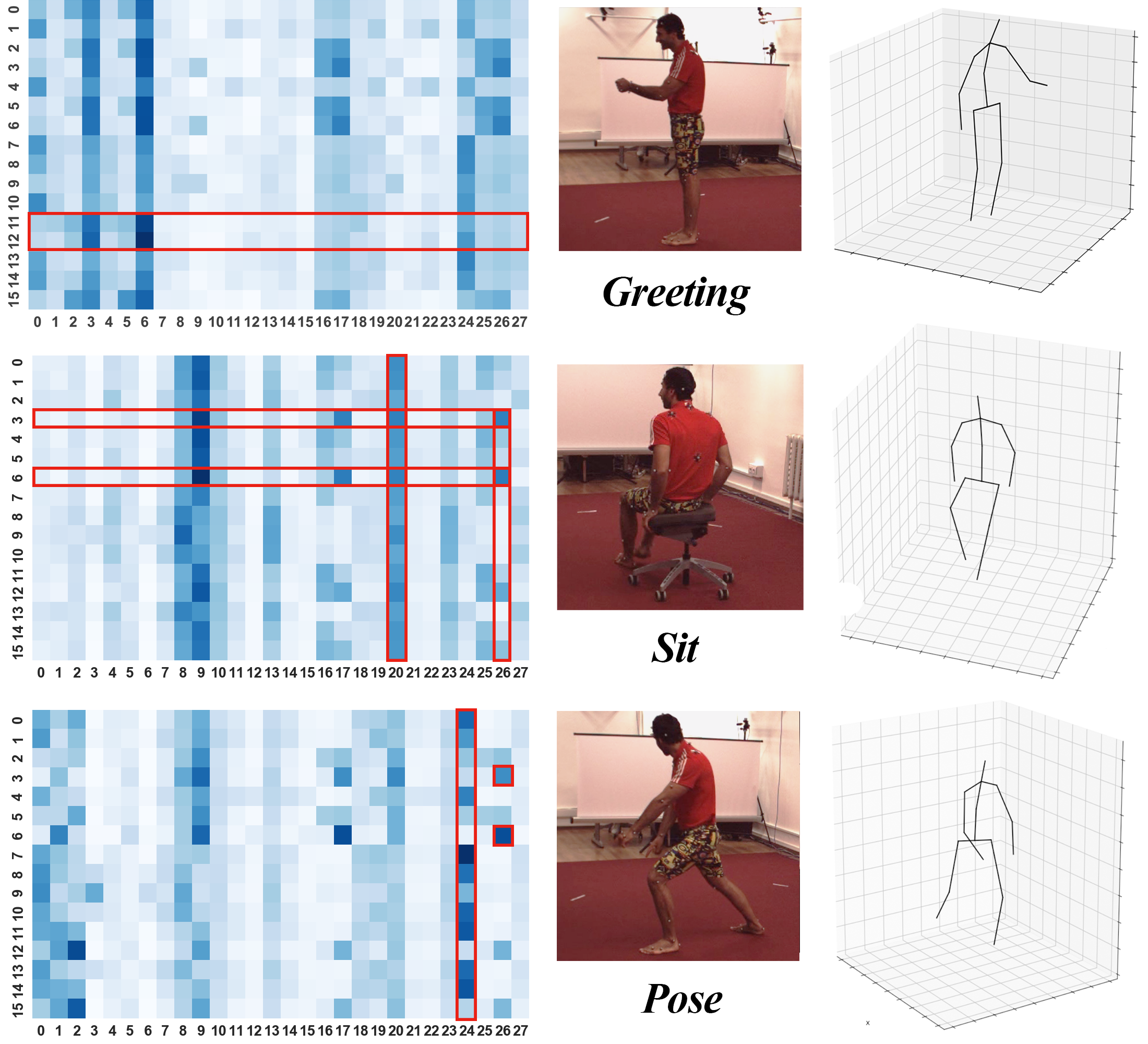}
	\caption{Visualization of attention maps in different actions. The proposed method has the ability to model motion coordination under different actions. For example, the relations between the left hand (Joints 11 and 12) and the feets (Joints 3 and 6) in 'Greeting'.}
	\label{fig:VisualizationAttentionMapDifferentAction}
\end{figure}

As shown in Figure \ref{fig:VisualizationAttentionMapDifferentAction}, we supplement visual experiments under various actions. Evidently, our method effectively captures long-range dependencies across different actions. In the 'Sit', it learns an important {\it head part} at 20-th node, and the 26-th {\it hand region} constrains the 3rd and 6th {\it foot} joints. The same applies to 'pose' as well. {The PGA module adjusts the 3D positions of various joints by learning a key lower body region (Joints 24). The long-range relations between the right knee (Joint 2) and both hands (Joint 12 and 15) is also established at the keypoint-level.} This also indicates that our method is capable of learning the underlying long-range constraints of different behaviors.  Such grouping constraints can exert more effective constraints on the nodes at the extremities of the human skeleton.

\begin{figure*}[htb]
	\centering
	\includegraphics[width=\textwidth]{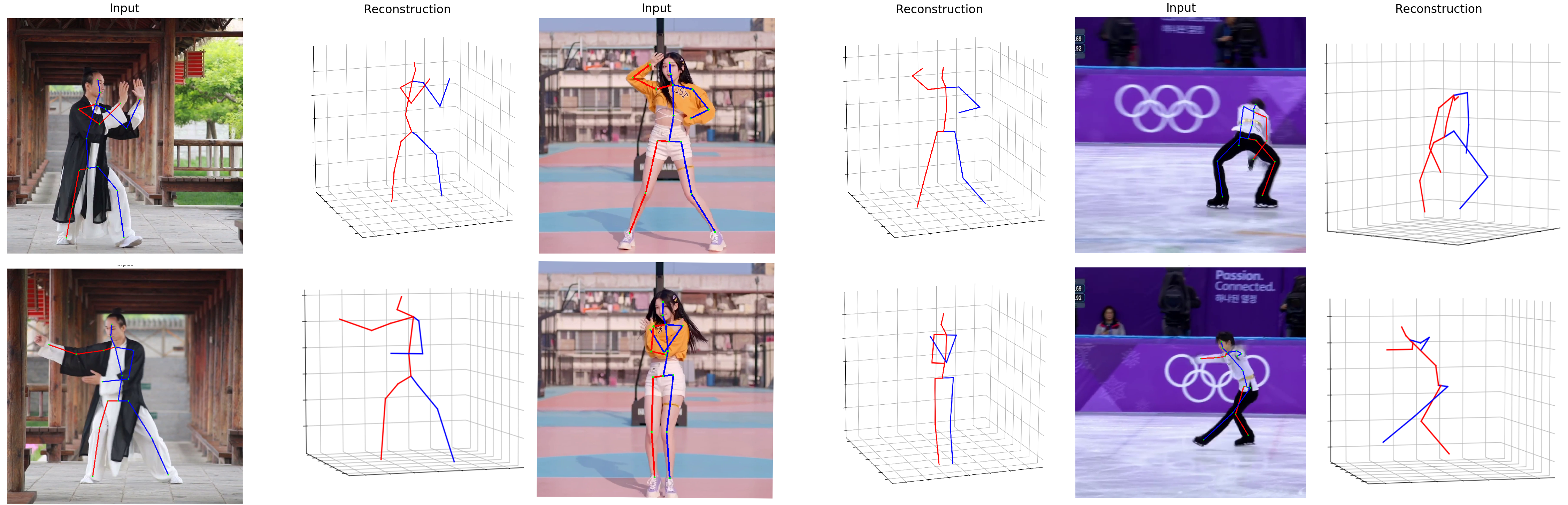}
	\caption{Visualization on various in-the-wild scenarios. These challenge scenarios of varying difficulty demonstrate the effectiveness and robustness of our approach. For example, the picture on the top right represents a severe self-occlusion action.}
	\label{fig:Visualization_in-the-wild}
\end{figure*}

To better showcase our real-world performance in Figure \ref{fig:Visualization_in-the-wild}, we have supplemented visualizations in some in-the-wild scenarios. From the qualitative results, our method, although only trained on Human3.6M, can effectively generalize to unseen scenes and actions. {In addition, it can be seen that our approach can be applied to different scenarios, including outdoor, sports, dance, etc.}

\section{Limitation}
In this work, we discover that long-range dependencies strongly correlate with diverse actions, especially complex ones. While we didn’t use text to model relationships between actions and poses, leveraging text for learning pose structures could be beneficial. In our proposed PGFormer and DiffPyramid, both versions balance accuracy and parameter count,  but neither achieves optimal performance in both. Additionally, the GMM processing format in DiffPyramid,  along with inference speed, further limits its application. Finally, single-frame methods, while simpler and efficient for spatial constraints, lack temporal smoothness and reduce accuracy. The impact of spatial long-range dependencies on temporal sequences will be explored in future.
\section{Conclusion}
\label{sec:Conclusion}
In this paper, we construct a pyramid connection structure from a multi-scale perspective to address challenges in 3D HPE. We propose Pyramid Graph Transformer (PGFormer), which is a simple and lightweight multi-scale transformer architecture. Our method encapsulates human sub-structure into Pyramid Graph Attention (PGA) module, which learns pyramid-structured long-range dependence efficiently through cross-scale correlation. In addition, the proposed PGFormer and PGA module are suitable for the architecture of the diffusion model (i.e., DiffPyramid). From extensive ablation studies and benchmark experiments, we can draw the conclusion: learning long-range dependencies reduces the errors of 3D HPE and exhibits robustness in complex actions. Moreover, long-range dependencies can be easily learned via cross-scale sub-graph information. Finally, we discuss the limitations of the proposed method and future work in this paper. We believe that further investigation into the relationship between spatial-temporal human constraints and text is meaningful. We hope it provides new insights in the field of 3D HPE.

\bibliographystyle{IEEEtran}
\bibliography{RefMain}

\begin{IEEEbiography}[{\includegraphics[width=1in,height=1.25in,clip,keepaspectratio]{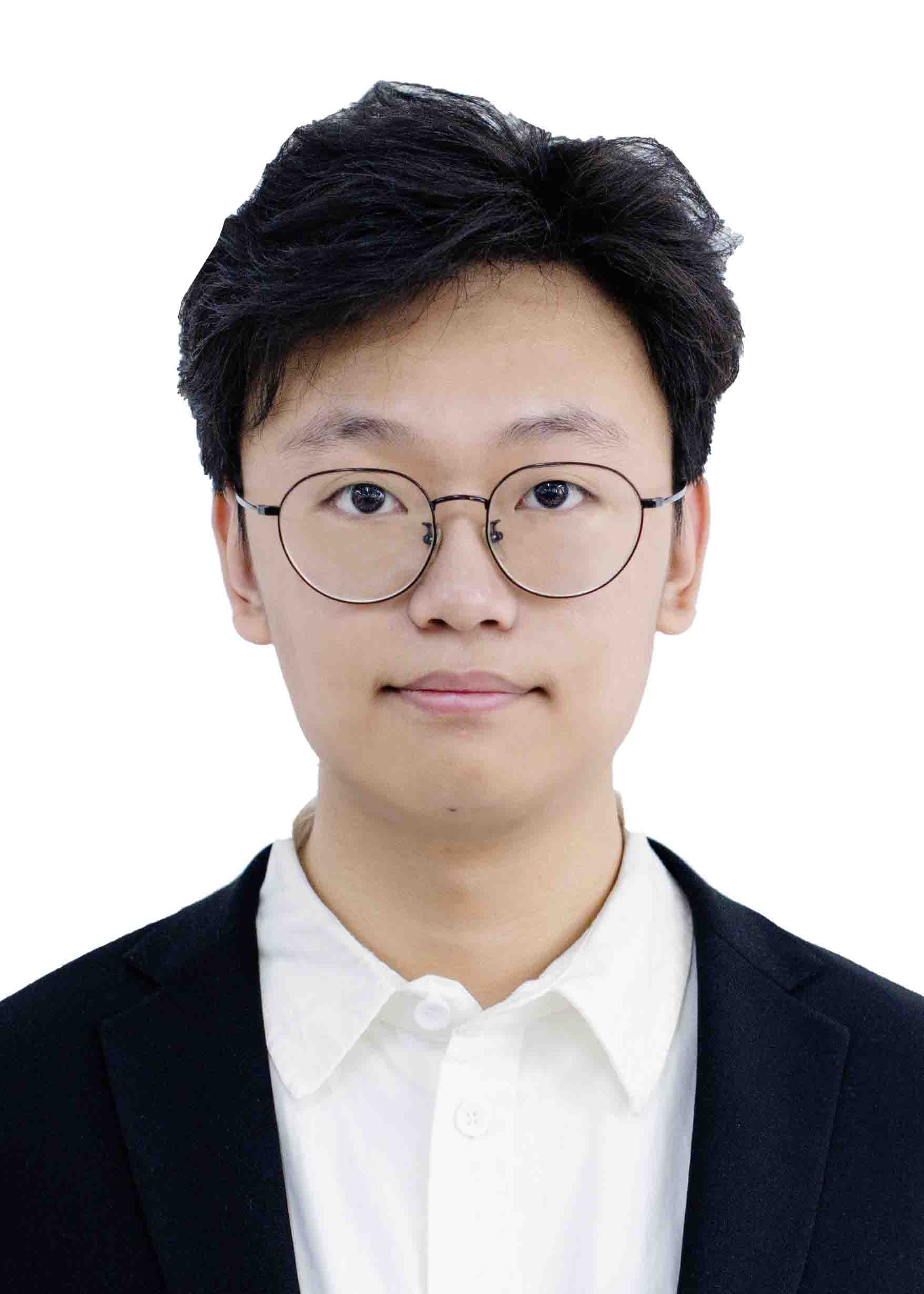}}]{Mingjie Wei} received the B.E. degree from China University of Geosciences, Wuhan, China, in 2023. He is currently pursuing an M.S. degree at Xidian University, Xi’an, China. His research interests include 3D human pose estimation,  motion generation and multimodal large language model.
\end{IEEEbiography}
\begin{IEEEbiography}[{\includegraphics[width=1in,height=1.25in,clip,keepaspectratio]{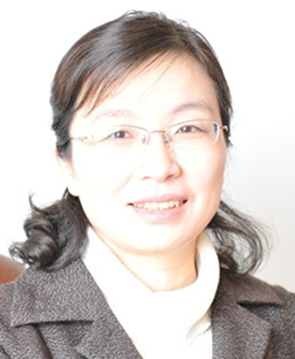}}]{Xuemei Xie} (Senior Member, IEEE) received the M.S. degree in electronic engineering from Xidian University, Xi’an, China, in 1994, and the Ph.D. degree in electrical and electronic engineering from The University of Hong Kong, Hong Kong, in 2004. She is currently a Professor with the School of Artificial Intelligence, Xidian University. She has authored more than 100 academic papers in international and national journals, and international conferences. Her research interests include scene understanding, embodied intelligence robotics, natural language processing, vision-language multimodal learning, and human–computer interaction.
\end{IEEEbiography}
\begin{IEEEbiography}[{\includegraphics[width=1in,height=1.25in,clip,keepaspectratio]{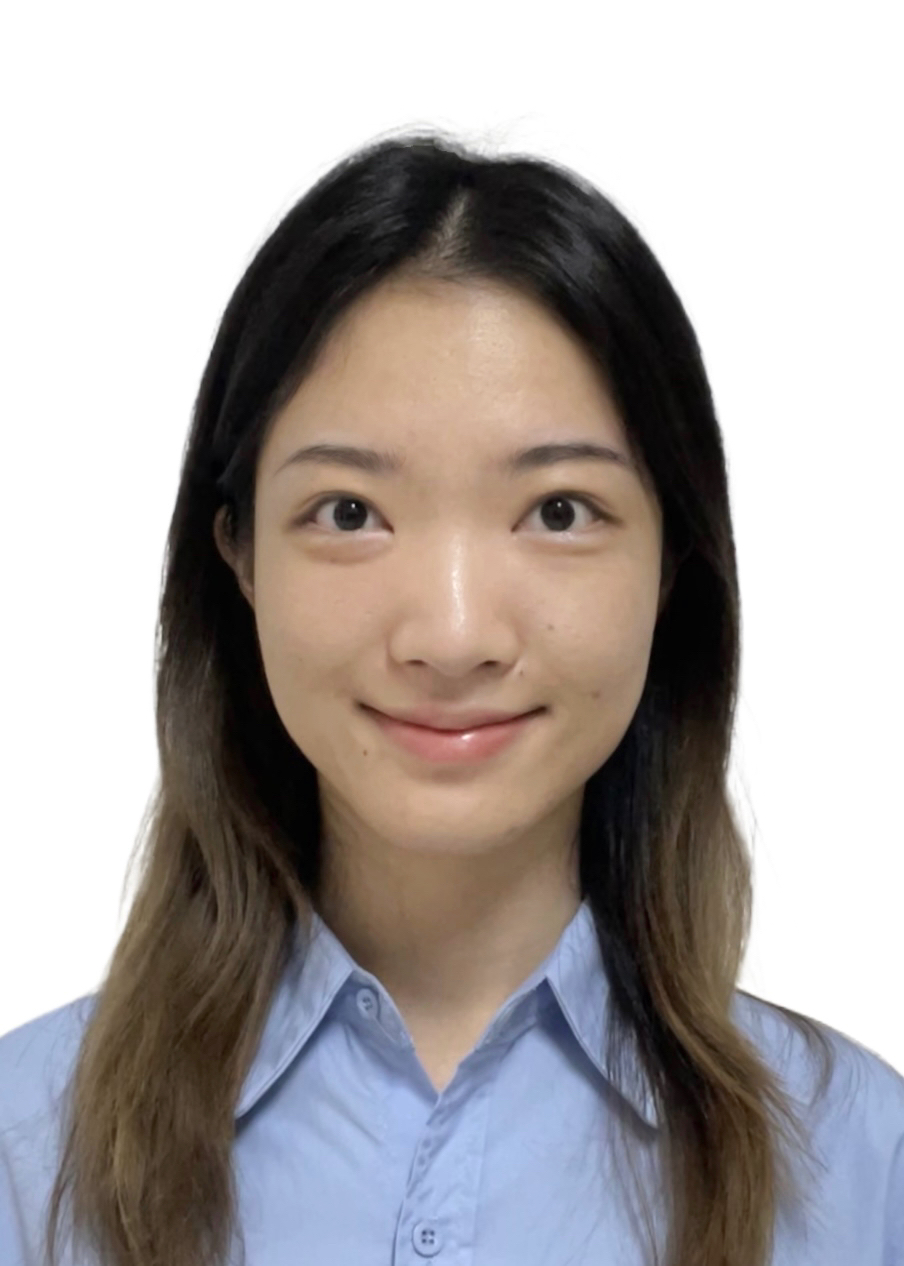}}]{Yutong Zhong} received the B.E. degree from Shaanxi normal university, Xian, China, in 2022. She is currently pursuing an M.S. degree at Xidian University, Xi’an, China. Her research interests include action recognition, temporal action segmentation and multimodal large language model.
\end{IEEEbiography}
\begin{IEEEbiography}[{\includegraphics[width=1in,height=1.25in,clip,keepaspectratio]{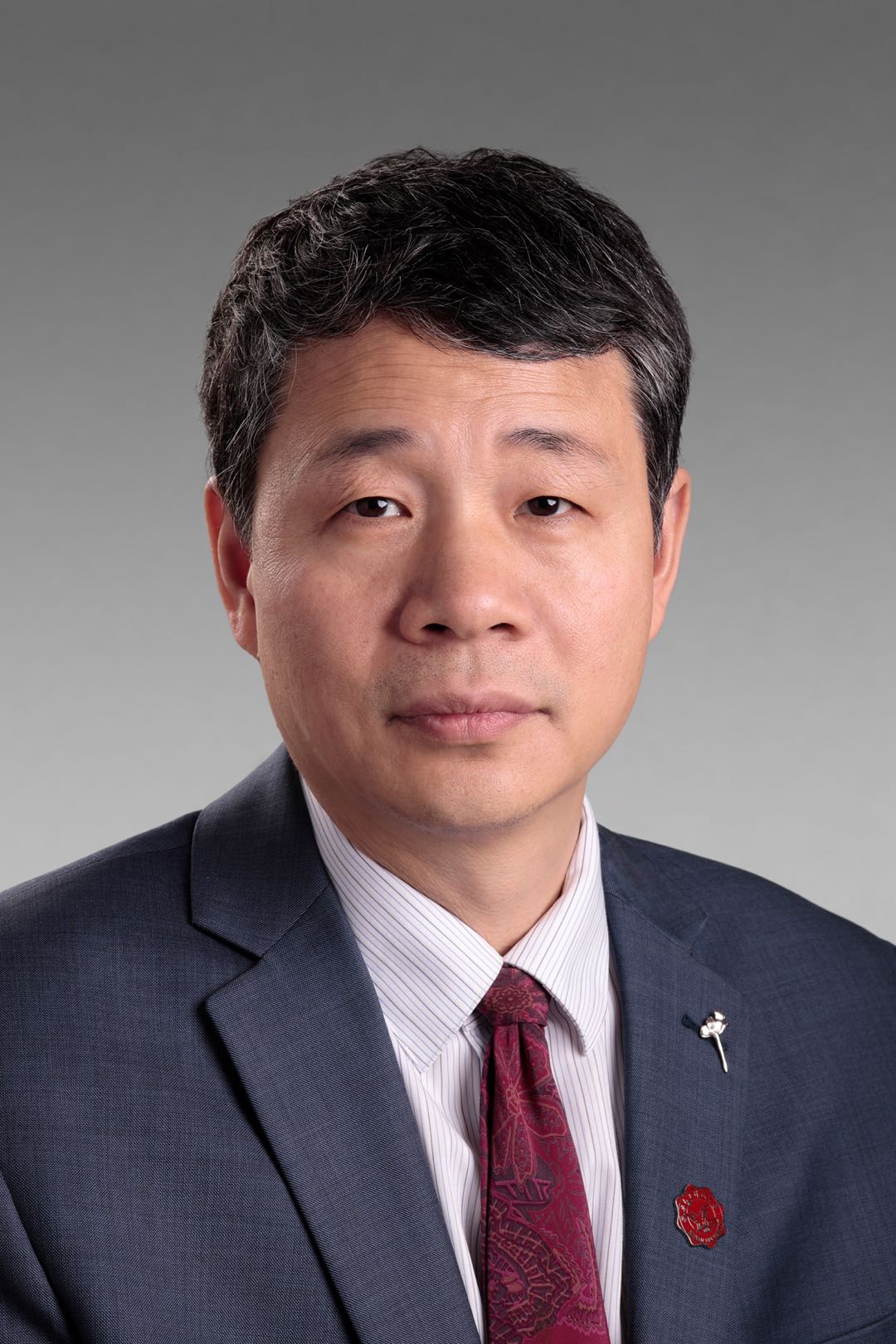}}]{Guangming Shi} (Fellow, IEEE) received the M.S. degree in computer control, and the Ph.D. degree in electronic information technology from Xidian University, Xi’an, China, in 1988, and 2002, respectively. His research interest includes artificial intelligence, semantic signal processing and communication, embodied intelligence robotics, and human-computer interaction. He is a Professor with the School of Artificial Intelligence, Xidian University. He is an IEEE Fellow and the chair of IEEE CASS Xi’an Chapter, Fellow of Chinese Institute of Electronics. He was awarded Cheung Kong scholar Chair Professor by the ministry of education in 2012 and ACM China Distinguished Researcher in 2024. And he was awarded with the Second Prize of China's National Natural Science Award in 2017.
\end{IEEEbiography}

\end{document}


\twocolumn[
\begin{center}
	\textbf{\Large Supplementary Materials for "Learning Pyramid-structured Long-range Dependencies for 3D Human Pose Estimation"}
\end{center}
\vspace{1em}
]
\section{Analysis of complexity}
Let the number of joints be denoted as $N$, the dimension of the hidden layer as $d_k$, and the $i$-th dimension of pooling from 0 to $n$ as $M_i$. Moreover, the number of nodes $N$ used for 3D human pose estimation represents the human body sparsely, typically limited to 16 or 17. After several pooling operations, $M_i$ is expected to fall within the range of 1 to 8. Additionally, $d_k$ is often significantly larger than both $N$ and $M_i$. Then, the time and space complexity required by vanilla MHSA is: $O(N^2d_k+Nd_k^2)$. In our approach, the time and space complexity required for pooling is linearly negligible $O(NC)$. Hence, the computational complexity of PGA is:
\begin{equation}
	O(PGA) = (N + \sum\nolimits_{i=0}^nM_i)^2d_k + (N + \sum\nolimits_{i=0}^nM_i)d_k^2.
\end{equation}
\par
So the proposed attention in PGA slightly increases time and space overhead, but not substantially. It reduces the overhead of GCN and MLP layers at each scale by integrating the multi-scale approach into the attention.\par
{\it Compare other multi-scale architectures.} Since d is much larger than N and M, the complexity of simplifying a GCN, GAT, or self-attention layer is $|V|d^2$, where $|V|$ is the number of keypoints computed at that layer. For GraphSH\cite{xu2021graph}, the feature dimension is increased after pooling, and each block is stacked with 5 layers. Its complexity is:
\begin{equation}
	O(GraphSH) = [2Nd_1^2+2M_1d_2^2+M_2d_3^3]*l,
\end{equation}
where, $l$ represents the number of stacked blocks of the network, i.e., 'Numlayer' in Methods. And $N=2M_1=4M_2, d_3=1.5d_2=2d_1$.For PHGANet\cite{zhang2023learning}, each of its modules is stacked with $s$ GAT blocks, and each block has a layer of self-attention and GCN, the complexity is:
\begin{equation}
	O(PHGANet) =  \sum\nolimits_{i=0}^{b_i-1} 2sN_id^2(l-b_i)
\end{equation}
where $b_i$ represents the number of branches For the proposed PGFormer, pooling is applied to self-attention, the complexity is:
\begin{equation}
	O(PGFormer) = [Nd^2+(N + \sum\nolimits_{i=0}^nM_i) d^2]*l.
\end{equation}
\par

\begin{table}[htb]
	\footnotesize
	\centering
	\caption{Comparison of complexity with other multi-scale methods. }
	\resizebox{0.7\columnwidth}{!}{
		\begin{tabular}{cccc}
			\toprule
			Multi-scale Method & Complexity  \\ 
			\midrule
			GraphSH \cite{xu2021graph} & $21Nd^2$ \\
			\rowcolor{mygray} Parallel multi-scale \cite{zhang2023learning} & $48Nd^2$ \\
			Ours(PGFormer) & $11Nd^2$ \\
			\bottomrule
		\end{tabular}
	}
	\label{table:complexity}
\end{table}

Table \ref{table:complexity} presents the final complexity result, with the pooled node
$M$ and dimension $d$ substituted equivalently. The result specifies 4 layers and 3 branches. The result of parameter number is consistent with SemGCN experiment.
\section{Additional Implementation deltails}
Our methods is implemented in PyTorch via Adam optimizer. PGFormer is conducted on a single NVIDIA RTX Titan GPU. In all experiments, we set the dropout rate to 0.25 after graph convolution and self-attention in the PGA module, and use 4 attention heads with a 0.05 dropout after softmax. $\lambda$ is set to 0 for the ablation study. For a fair comparison, we use the 17 keypoints CPN results, pooling the extra 'Nose' point with other head-related keypoints to form 8 part-level nodes. And no additional pose refinement \cite{cai2019exploiting, zou2021modulated} model is used in data augmentation.\par 
In DiffPyramid, we utilize 4 attention heads, with a subsequent dropout of 0.15 after the softmax calculation. The proposed DiffPyramid can be trained on a single NVIDIA RTX A6000 GPU within 24 hours.
\section{Visualization of attention map with $l1-norm$.}

\begin{figure}[htb]
	\centering
	\begin{minipage}{1\linewidth}
		\centering
		\centerline{\includegraphics[width=\textwidth]{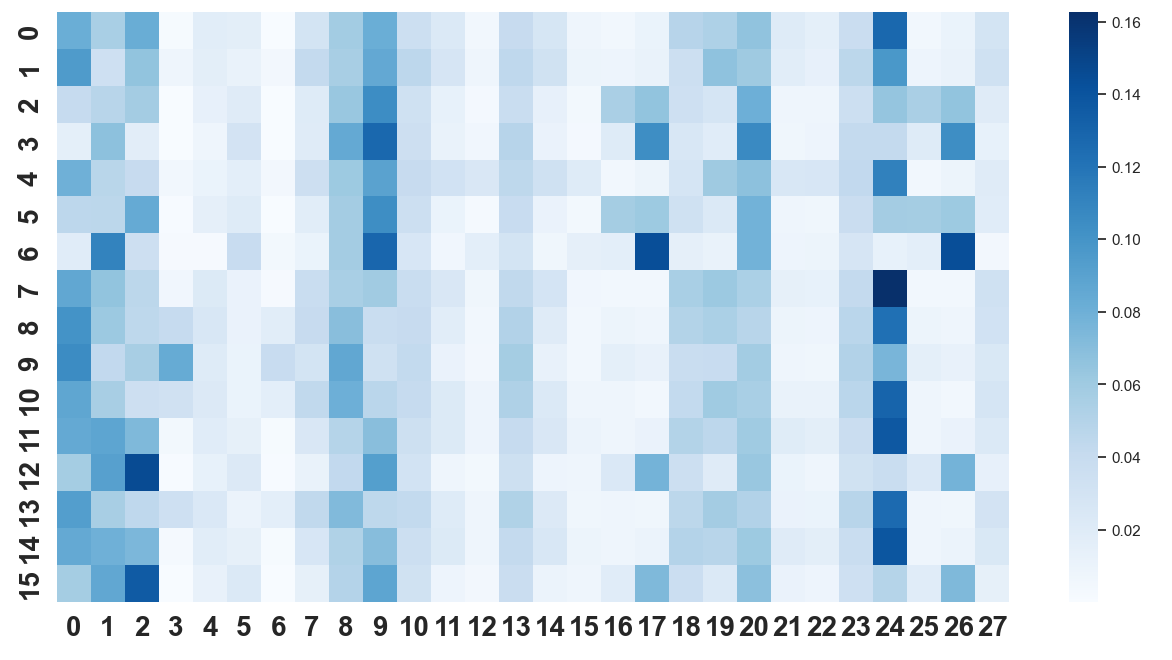}}
	\end{minipage}
	\centering
	\begin{minipage}[b]{1\linewidth}
		\centering
		\centerline{\includegraphics[width=\textwidth]{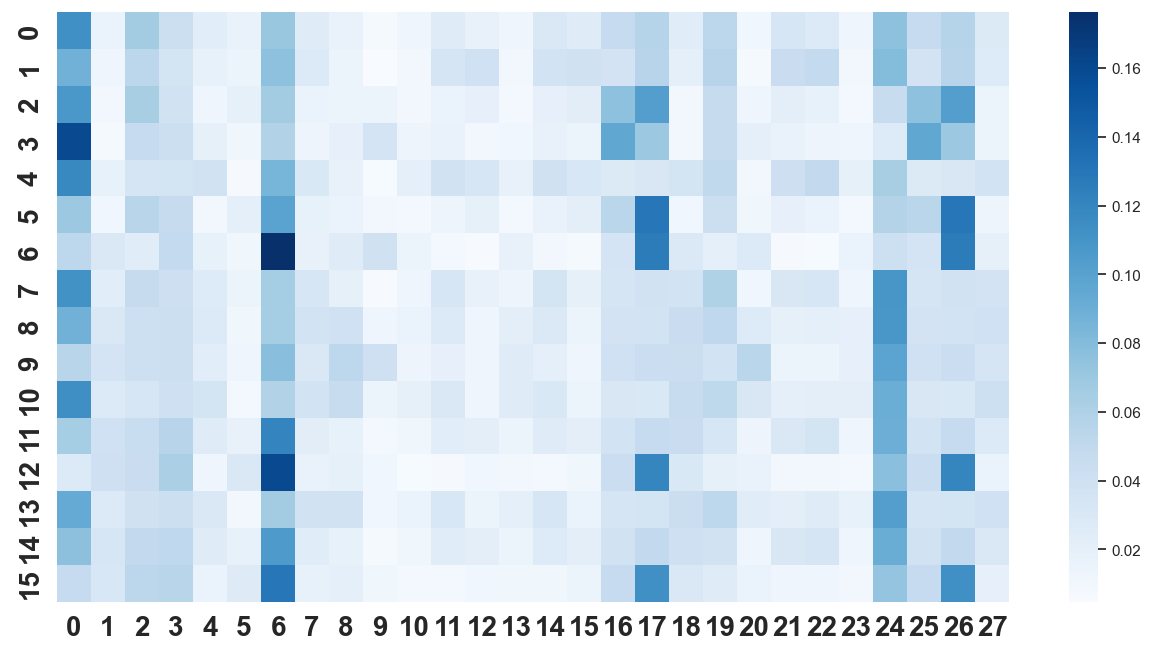}}
	\end{minipage}
	\caption{{\bf Top:} Diagram illustrating long-range dependencies learned by a PGA module without $l1-norm$. {\bf Bottom:} Corresponding diagram showing long-range dependencies learned by a PGA module with $l1-norm$. }
	\label{fig:l1_normAttentionMap}
\end{figure}

As shown in Fig. \ref{fig:l1_normAttentionMap}, the most intuitive effect of using $l1-norm$ is that the matrix becomes sparse, which makes some weakly correlated long-range information closer to 0. Second, this enhances important nodes (such as 0,6,17,24, and 26). So PGA module learns more efficient long distance constraints.

\section{A diagram of 17 keypoints group}
\begin{figure}[htb]
	\begin{minipage}[b]{1\linewidth}
		\centering
		\centerline{\includegraphics[width=6cm]{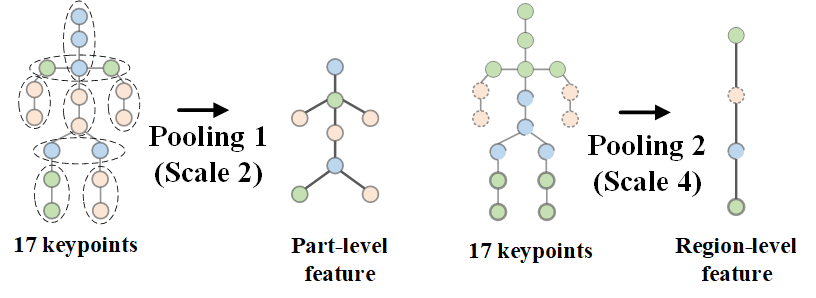}}
		\medskip
	\end{minipage}
	\caption{A diagram illustrating the pooling of 17 keypoints. The 'Nose' keypoints are grouped with other head-related nodes.}
	\label{fig:otherpooling17pt}
\end{figure}

As shown in Fig. \ref{fig:otherpooling17pt}, we implement the pooling scheme for 17 keypoints on the detection data.

\bibliographystyle{IEEEtran}
\bibliography{RefMain}